\documentclass{article} %
\usepackage{iclr2026_conference,times}

\iclrfinalcopy

\usepackage{amsmath}
\usepackage{amssymb}
\usepackage{amsfonts}
\usepackage{algorithmic}
\usepackage{graphicx}
\usepackage{textcomp}
\usepackage{xcolor}

\usepackage{indentfirst}
\usepackage{color}
\usepackage{enumitem}
\usepackage{multirow}
\usepackage{colortbl}
\usepackage{tabularray}
\usepackage{graphicx}
\usepackage{subfigure}
\usepackage{graphicx, nicefrac}
\usepackage{url}
\usepackage[flushleft]{threeparttable}
\usepackage{bigstrut}
\usepackage{ifthen}

\usepackage{makecell}
\usepackage{diagbox}
\usepackage{soul}
\usepackage[most]{tcolorbox}
\usepackage{colortbl}

\usepackage{ulem}
\usepackage{makecell} %
\usepackage{booktabs} %
\usepackage{caption}
\usepackage[hyperfootnotes=false]{hyperref}
\usepackage{soul,color}
\usepackage{url}
\usepackage{graphicx}
\usepackage{xspace}
\usepackage{xcolor}
\usepackage{wrapfig}
\usepackage{lipsum}
\usepackage{longtable}
\usepackage{tikz}
\usetikzlibrary{tikzmark}
\usepackage{enumitem}
\setlist{leftmargin=5.5mm}

\usepackage{booktabs}
\usepackage{multirow}
\usepackage{threeparttable}
\usepackage{caption}
\usepackage{rotating}
\usepackage{subcaption}
\usepackage{colortbl}

\usepackage[capitalize,noabbrev]{cleveref}

\usepackage[ruled,linesnumbered]{algorithm2e}

\usepackage{amsmath,amsfonts,bm}

\def\eqref#1{equation~\ref{#1}}

\def\1{\bm{1}}

\def\vx{{\bm{x}}}

\def\mE{{\bm{E}}}

\def\mI{{\bm{I}}}

\def\mX{{\bm{X}}}

\DeclareMathAlphabet{\mathsfit}{\encodingdefault}{\sfdefault}{m}{sl}
\SetMathAlphabet{\mathsfit}{bold}{\encodingdefault}{\sfdefault}{bx}{n}

\newcommand{\eg}{{\it e.g.}}

\newcommand{\ie}{{\it i.e.}}
\newcommand{\etc}{{\it etc.}}

\newcommand{\model}{\textsc{VisionTS++}\xspace}
\newcommand{\visionts}{\textsc{VisionTS}\xspace}
\newcommand{\mae}{\texttt{MAE}\xspace}

\definecolor{fbApp}{HTML}{ffe4e3}
\definecolor{tabhighlight}{HTML}{e5e5e5}
\newcommand{\rowc}{\rowcolor{fbApp}}

\tikzset{mycircled/.style={circle,draw,inner sep=0.05em,line width=0.04em, scale=0.8}}

\definecolor{pink}{rgb}{1, 0, 0.5}
\definecolor{darkgrey}{rgb}{0.53,0.53,0.53}
\definecolor{mygrey}{rgb}{0.9,0.9,0.9}
\definecolor{purple}{RGB}{230, 227, 254}
\definecolor{lightgreen}{RGB}{238, 252, 241}
\definecolor{lightred}{RGB}{231, 187, 187}
\definecolor{darkred}{RGB}{198, 129, 129}

\definecolor{tabhighlight}{HTML}{e5e5e5}
\definecolor{someorange}{rgb}{0.773,0.353,0.067}
\definecolor{someblue}{rgb}{0.27, 0.35, 0.760}
\definecolor{codegreen}{rgb}{0,0.5,0}
\definecolor{codeblue}{rgb}{0.25,0.5,0.5}
\definecolor{codegray}{rgb}{0.6,0.6,0.6}

\newcommand{\emoji}[2][1.2em]{\raisebox{-0.2\height}{\includegraphics[height=#1]{#2}}}
\definecolor{fbApp}{HTML}{ffe4e3}
\definecolor{mydarkblue}{rgb}{0,0.3,0.9}

\hypersetup{colorlinks=true,citecolor=mydarkblue}

\newcommand{\moirai}{\textsc{Moirai}\xspace}

\usepackage[utf8]{inputenc} %
\usepackage[T1]{fontenc}    %
\usepackage{url}            %
\usepackage{booktabs}       %
\usepackage{amsfonts}       %
\usepackage{nicefrac}       %
\usepackage{microtype}      %
\usepackage{xcolor}         %

\title{$\vcenter{\hbox{\includegraphics[width=2.5ex,height=2.5ex]{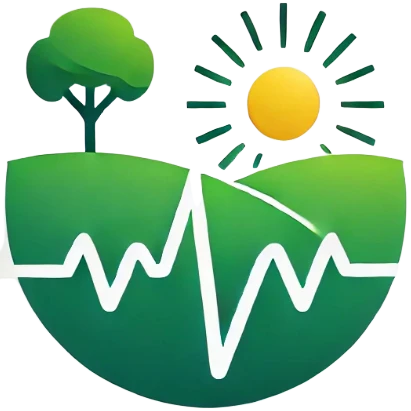}}}$ VisionTS++: Cross-Modal Time Series Foundation Model with Continual Pre-trained Vision Backbones}

\author{
\textbf{Lefei Shen}$^{1,*}$, \textbf{Mouxiang Chen}$^{1,}$\thanks{Both authors contributed equally to this research. $^\dagger$Corresponding authors.}~~, \textbf{Xu Liu}$^2$, \textbf{Han Fu}$^1$, \\
\textbf{Xiaoxue Ren}$^{1}$, \textbf{Jianling Sun}$^{1}$, 
\textbf{Zhuo Li}$^{3,\dagger}$, \textbf{Chenghao Liu}$^{4,\dagger}$
\\
$^1$ Zhejiang University \quad 
$^2$ National University of Singapore\\
$^3$ State Street Technology (Zhejiang) Ltd. \quad
$^4$ Salesforce Research Asia\\
\texttt{\{lefeishen, chenmx, 11821003, xxren, sunjl, lizhuo\}@zju.edu.cn} \\
\texttt{\{liuxu726, twinsken\}@gmail.com}
}

\begin{document}

\maketitle

\begin{abstract}

Recent studies have indicated that vision models pre-trained on images can serve as time series foundation models (TSFMs) by reformulating time series forecasting (TSF) as image reconstruction. However, effective cross-modal transfer from vision to time series remains challenging due to three discrepancies: (1) the \textbf{data-modality gap} between structured, bounded image data and unbounded, heterogeneous time series; (2) the \textbf{multivariate-forecasting gap} between fixed RGB-three-channel vision models and time series with arbitrary numbers of variates; and (3) the \textbf{probabilistic-forecasting gap} between the deterministic outputs of vision models and the requirement for uncertainty-aware probabilistic predictions. 
To bridge these gaps, we propose \model, a TSFM based on continual pre-training of a vision model on large-scale time series.  Our approach introduces three key innovations: (1) \textbf{vision-model-based filtering} to identify high-quality sequences to stabilize pre-training and mitigate modality gap; (2) \textbf{colorized multivariate conversion}, encoding multivariate series as multi-subfigure RGB images to enhance cross-variate modeling; (3) \textbf{multi-quantile forecasting}, using parallel reconstruction heads to generate quantile forecasts without parametric assumptions. 
Experiments show that \model achieves state-of-the-art performance in both in-distribution and out-of-distribution forecasting, outperforming specialized TSFMs by 6\%-44\% in MSE reduction and ranking first in GIFT-Eval benchmark which comprises 23 datasets across 7 domains. Our work demonstrates that with appropriate adaptation, vision models can effectively generalize to TSF, thus advancing the pursuit of universal TSFMs. 
Code is available at \url{https://github.com/HALF111/VisionTSpp}.

\end{abstract}

\section{Introduction} \label{introduction}

\par Foundation models have transformed natural language processing (NLP) \citep{BERT, GPT-2} and computer vision (CV) \citep{ViT, MAE, Swin_transformer}, motivating the development of \textit{time series foundation models} (TSFMs) for \textit{universal forecasting}—i.e., a single model that generalizes across diverse tasks without task-specific training \citep{Moirai, Chronos, TimesFM, Time-MoE}. 
Yet, the heterogeneity of time series—spanning scale, frequency, and dimensionality—poses a major challenge to unified modeling \citep{Timer, Chronos, Moirai-MoE}.

\par Recent work suggests that vision models pre-trained on images can be surprisingly effective for time series forecasting (TSF) \citep{DMMV, vision4ts_survey, vision4ts_survey_4}. 
Notably, \citet{VisionTS} shows that by reformulating univariate forecasting as image reconstruction, a Masked Autoencoder (\mae) pre-trained on natural images matches or exceeds specialized TSFMs. 
This hints at a conceptual alignment: images and time series may share similar patterns—\eg, textures and edges in images can correspond to periodicities and trends in time series.

\par However, despite this promise, some fundamental discrepancies between them limit further improvements. Specifically, we identify three key gaps: 
\textbf{Data-Modality Gap}: Image pixels are bounded and spatially structured; while time series are unbounded and temporally heterogeneous. Directly applying vision models to TSF without appropriate adaptation is therefore suboptimal. 
\textbf{Multivariate-Forecasting Gap}: Vision models are designed with fixed RGB-three-channels, while multivariate time series have arbitrary numbers of variates (also referred to as \textit{channels} in this paper). This hinders effective modeling of cross-variate dependencies.
\textbf{Probabilistic-Forecasting Gap}: Most vision models focus on deterministic tasks like reconstruction, yet practical TSFMs require effective uncertainty-aware probabilistic predictions \citep{Moirai, Chronos, Moirai-MoE}.

\par A straightforward yet blunt approach involves architectural modifications—\eg, replacing input/output layers with time-series-specific modules \citep{GPT4TS, Time-LLM, Chronos}—followed by continual pre-training (CPT). 
However, such changes can disrupt valuable pre-trained visual representations, leading to negative transfer \citep{VisionTS, negative_transfer_1, negative_transfer_2}, and further degrade performance due to noisy or low-quality time series data \citep{Timer, Chronos}. For example, \citet{GPT4TS} observe poor results when directly fine-tuning BeiT \citep{BeiT} for forecasting.
This raises a critical question: \textbf{How can we effectively adapt a pre-trained vision model for TSF tasks, maximizing transfer effectiveness while robustly preserving its original knowledge?}

\begin{figure*}[t]
    \centering  %
    \includegraphics[width=\textwidth]{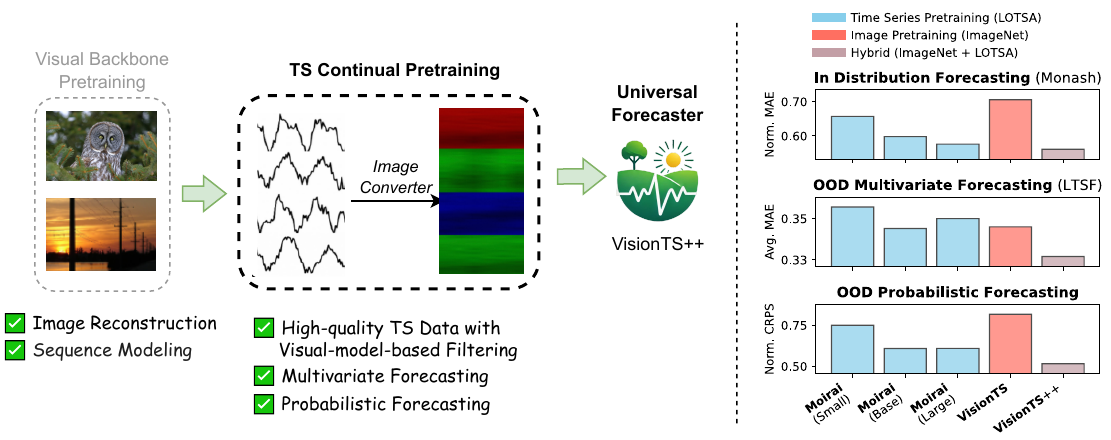}
    \caption{\textbf{Left}: Training pipeline of \model. We perform continual pre-training of a visual backbone (\mae) on large-scale time series datasets to create a powerful and universal TSFM.
    \textbf{Right}: \model outperforms \moirai and \visionts in both multivariate and probabilistic forecasting, demonstrating its superior effectiveness.}
    \vspace{-1.3em}
    \label{fig:teaser}
\end{figure*}

Building upon the framework of \visionts, our philosophy is to minimally modify the \mae architecture, and also transform TSF into image reconstruction. 
Based on this, we propose \model, a vision-model-based TSFM that undergoes continual pre-training on large-scale time series, which supports flexible multivariate and probabilistic forecasting by efficiently transferring visual knowledge for TSF.
Specifically, \model includes three key innovations to bridge the above gaps:

\begin{itemize}
    \item \textbf{Vision-Model-Based Filtering}: To address the \textit{data-modality gap}, we introduce a filtering mechanism that leverages the vision model itself to select high-quality time series. we identify and discard samples with out-of-range values or abrupt anomalies—inputs incompatible with the model’s constraints. This enhances pre-training stability and mitigates negative transfer.
    \item \textbf{Colorized Multivariate Conversion}: To handle the \textit{multivariate-forecasting gap}, we encode multivariate time series as multi-subfigure RGB images, where each variate is mapped to a distinct subfigure. This allows cross-variate dependencies to be better captured as spatial relationships between subfigures—naturally aligning with \mae’s multi-object analysis capability.
    \item \textbf{Multi-Quantile Forecasting}: To tackle the \textit{probabilistic-forecasting gap}, we employ parallel reconstruction heads that generate multiple output images, each corresponding to a different quantile forecast. This reformulates probabilistic prediction as a set of deterministic image reconstructions—enabling flexible, assumption-free distribution modeling without relying on parametric priors \citep{Moirai}.
\end{itemize}

\par After continual pre-training with these adaptations, \model achieves state-of-the-art (SOTA) performance across diverse forecasting tasks.
For in-distribution forecasting, \model achieves the best normalized MAE on the Monash benchmark \citep{Monash}.
For out-of-distribution evaluations, \model outperforms existing TSFMs by 6\%–44\% in MSE reduction on the long-term TSF benchmark \citep{Autoformer}. It also ranks first in the Probabilistic Forecasting benchmark \citep{Moirai} and GIFT-Eval benchmark \citep{gift-eval} which comprises 23 datasets across 7 domains, beating many specialized TSFMs, thus demonstrating its strong generalization ability.

\par The training pipeline of \model is summarized in \cref{fig:teaser}. And our key contributions are summarized as follows:
\begin{itemize}
    \item We propose \model, a novel TSFM that performs continual pre-training of vision models on large-scale time series datasets, effectively adapting the model to time series temporal patterns while preserving pre-trained visual knowledge.
    \item We propose three targeted innovations—vision-model-based filtering, colorized multivariate time series conversion, and multi-quantile forecasting—that systematically address the data-modality, multivariate-forecasting, and probabilistic-forecasting gaps in cross-modal transfer.
    \item We demonstrate SOTA performance across in-distribution (\eg, Monash) and out-of-distribution (\eg, LTSF, PF, GIFT-Eval) benchmarks, establishing \model as a robust and general-purpose TSFM.
\end{itemize}

\section{Preliminaries}

\paragraph{Time Series Forecasting (TSF)}
\par For a multivariate time series with $M$ \textit{variates} (also referred to as \textit{channels} in this paper), let $\vx_{t} \in \mathbb{R}^M$ represent the value at $t$-th time step.
Then given a historical sequence (\ie, look-back window) $\mX_{t-L:t} = [\vx_{t-L}, \cdots, \vx_{t-1}] \in \mathbb{R}^{L\times M}$ with a context length of ~$L$, the TSF task is to use $\mX_{t-L:t}$ to predict future values (\ie, forecasting window): $\hat{\mX}_{t:t+T} = [\hat\vx_{t}, \cdots, \hat\vx_{t+T-1}] \in \mathbb{R}^{T\times M}$, where $T$ is the prediction length.

\paragraph{Image Reconstruction Task in \mae}
The Masked Autoencoder (\mae)~\citep{MAE} learns visual representations by reconstructing masked patches of an image. Given a square image of size $W \times W$, it is divided into $N \times N$ patches, each with a width and height of $S = \nicefrac{W}{N}$. During pre-training, random patches are masked, and a Vision Transformer (ViT)~\citep{ViT} is trained to reconstruct the missing pixel values based on the visible patches.

\paragraph{Quick Review of \visionts}
Before introducing \model, we briefly revisit the \visionts model \citep{VisionTS}. Its core idea is to reformulate TSF as an image reconstruction task to adapt \mae for forecasting, which involves five key steps:
(1) Segmentation and Image Conversion: It first segments a 1D time series $\vx \in \mathbb{R}^L$ into periodic subsequences of length $P$, then arranges them into a 2D matrix $\mI_{\text{raw}} \in \mathbb{R}^{P \times \lfloor L/P \rfloor}$. (2) Normalization and Rendering: After the instance normalization which yields $\mI_{\text{norm}}$, the matrix is rendered into a grayscale-like image by repeating values across three RGB channels. 
(4) Alignment: To align with \mae’s input format, the image is resized to $(N\cdot S) \times (n\cdot S)$, where $n = \lfloor N \cdot L / (L+T) \rfloor$, so that the left portion corresponds to the context and the right portion (masked) to the forecast horizon. (5) Reconstruction and Time-series Conversion: The \mae model reconstructs the image, and the masked region is converted back to a 1D forecast through inverse operations.

\section{Methodology} \label{sec:methodology}

In this section we present \model, a vision-model-based TSFM that adapts the pre-trained \mae backbone via \textbf{Continual Pre-training} (CPT) on large-scale time series data, enabling the vision model to align with the patterns of time series data. Building on \visionts \citep{VisionTS}, we also reformulate TSF as an image reconstruction task. 
However, direct CPT is insufficient and faces three key challenges: the \textit{Data-Modality Gap}, \textit{Multivariate-Forecasting Gap}, and \textit{Probabilistic-Forecasting Gap}, which hinder effective cross-modal transfer between images and time series.
To bridge these gaps, we introduce three targeted designs—illustrated in Figure~\ref{fig:main_framework}—that require minimal architectural changes while significantly improving adaptation and generalization.

\subsection{Vision-Model-Based Filtering for Time Series Pre-training} \label{subsec:vision-model-based-filtering}

\par Firstly, to bridge the \textbf{Data-Modality Gap}, the core idea of \model is to perform continual pre-training (CPT) on large-scale time series data. 
However, the inherent heterogeneity and high noise in real-world time series raise concerns about data quality \citep{Timer, Chronos, Time-MoE}, thus demanding effective data curation approaches.

\par To obtain high-quality datasets, prior work in language models \citep{LLM_filter_1, LLM_filter_2, LLM_filter_3} and vision-language models \citep{VLM_filter_1, VLM_filter_2, VLM_filter_3} has demonstrated that data filtering strategies can significantly improve dataset quality. 
Inspired by them, we explore the feasibility of similar techniques for time series—but a key question arises: ``\textbf{How can we effectively filter low-quality time series to better bridge the data-modality gap for vision models?}''

\par
\begin{figure*}[t]
    \centering
    \includegraphics[width=\textwidth]{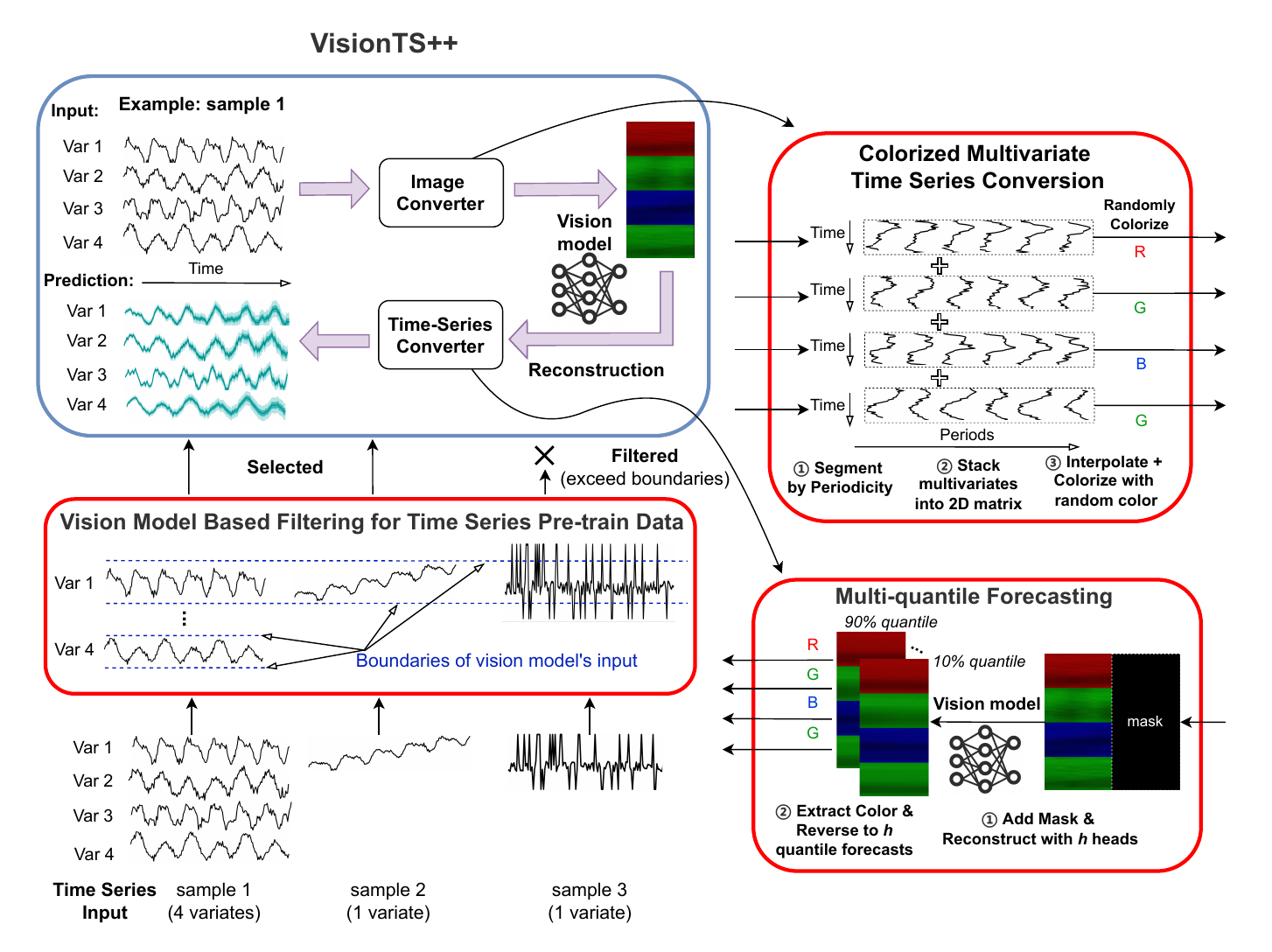}
    \vspace{-1.5em}
    \caption{ %
    Overview of \model. For each input, the following pipeline is applied: (1) Samples with out-of-range values after normalization are filtered out; (2) Each variate is segmented by periodicity and rendered as a colored subfigure, forming a composite image; (3) Multiple quantile forecasts are generated via parallel reconstruction heads. The model conducts continual pre-training on such transformed time series data to adapt \mae for universal forecasting.
    }
    \vspace{-1.2em}
    \label{fig:main_framework}
\end{figure*}

 To tackle this, we propose ``\textbf{Vision-Model-Based Filtering}'' (see bottom left part of Figure \ref{fig:main_framework}), which uses the vision model’s own input constraints as a criterion to identify and filter out low-quality time series. This is based on the observation that vision models expect inputs within a bounded range (\eg, image raw pixels in $[0, 255]$), whereas time series values are often unbounded. Time series containing out-of-range values can disrupt the model's pre-trained visual knowledge and harm transfer performance \citep{Timer, Chronos}. 

\par Specifically, pre-trained vision models expect inputs within a fixed range (between 0 and 255) derived from their training data (\eg, ImageNet). Given pixel values, after normalization using dataset mean $\mu_I$ and standard deviation $\sigma_I$, valid inputs lie within the interval: $\left[ \nicefrac{(0-\mI_\text{mean})}{\mI_\text{std}}, \nicefrac{(255 - \mI_\text{mean})}{\mI_\text{std}} \right]$.
Then for a time series input $\mX_{t-L:t} \in \mathbb{R}^{L\times M}$ and target $\hat{\mX}_{t:t+T} \in \mathbb{R}^{T\times M}$, we apply instance normalization using the context statistics $\mu_\mX = \mathrm{mean}(\mX_{t-L:t})$ and $\sigma_\mX = \mathrm{std}(\mX_{t-L:t})$. 

\par Furthermore, to align the dynamic range with that of images, we follow \visionts and scale the normalized values by a factor $r = 0.4$, obtaining: $\mX_{t-L:t}^{norm} = r \cdot \frac{\mX_{t-L:t} - \mu_\mX}{\sigma_\mX}$ and $\mX_{t:t+T}^{norm} = r \cdot \frac{\mX_{t:t+T} - \mu_\mX}{\sigma_\mX}$ for both input and target.
Despite this scaling, some values may still fall outside the valid visual input range. We thus filter out any sample for which $\mX_{t-L:t}^{norm}$ or $\hat{\mX}_{t:t+T}^{norm}$ contains values beyond $\left[ \nicefrac{(0-\mu_I)}{\sigma_I},\ \nicefrac{(255 - \mu_I)}{\sigma_I} \right]$, ensuring compatibility with the vision model’s input distribution.

\subsection{Colorized Multivariate Time Series Conversion}

\par Having filtered high-quality samples, we need an image converter to transform multivariate time series into 2D images for the vision backbone. While \visionts~\citep{VisionTS} processes each variate independently, this channel-wise isolation limits cross-variate modeling and increases computational overhead. A more scalable approach must support arbitrary numbers of variates within a unified visual representation. This leads to a critical question:
\textbf{“How can we extend the image-based approach to better support efficient and effective multivariate time series forecasting?”}

\par A straightforward solution is to utilize the RGB channels as carriers for the multiple variates. However, there exists a significant ``\textbf{Multivariate-Forecasting Gap}'' between them: standard vision models assume exactly three input channels, which cannot naturally accommodate the high dimension of time series with arbitrary numbers of variates.

\par To bridge this gap, we propose ``\textbf{Colorized Multivariate Conversion}'', which treats each variate as a distinct \textbf{subfigure} within a single composite image (see top right of Figure \ref{fig:main_framework}). Rather than using RGB channels to encode variate values, we use them to define \textbf{spatial boundaries}, enabling the vision model to leverage its native multi-object analysis capability for cross-variate dependency modeling.

\par Formally, for input $\mX_{t-L:t} \in \mathbb{R}^{L \times M}$, we follow \visionts to segment each variate into $\lfloor \nicefrac{L}{P} \rfloor$ patches of length $P$ (periodicity), reshaping into a $P \times \lfloor \nicefrac{L}{P} \rfloor$ matrix. This yields $\mI_{\text{raw}} \in \mathbb{R}^{M \times P \times \lfloor \nicefrac{L}{P} \rfloor}$. Each subfigure is then resampled to size $(\lfloor \nicefrac{W}{M} \rfloor, \nicefrac{W}{2})$, where $W$ is the image width. Notably, we fix the visible and masked regions each to a width $\nicefrac{W}{2}$, enabling efficient batch training across variable-length inputs.

\par Subsequently, $M$ subfigures are vertically stacked into a single image of size $(M \cdot \lfloor \nicefrac{W}{M} \rfloor, \nicefrac{W}{2})$ and placed on the left side of the image. In case $M$ is not evenly divided by $W$, 
zero-padding is applied at the bottom of images. This layout ensures all variates are processed jointly in one forward pass.

\par Furthermore, to enhance clear boundaries between variates, we assign each subfigure a random RGB channel (others zeroed), with adjacent subfigures guaranteed to use different channels. This \textbf{color-as-boundary} strategy serves three important purposes: (1) activates the vision model’s inherent multi-object capabilities; (2) prevents color bias through randomization; and (3) scales naturally to high-dimensional inputs.

\subsection{Multi-Quantile Forecasting for Probabilistic Conversion}

\par After image conversion and masking, the visual backbone reconstructs the right half of the image. While this supports point forecasting, standard vision models that are designed for deterministic tasks lack inherent mechanisms for uncertainty quantification, which is a key requirement in most TSFMs \citep{Moirai, Chronos}. We term such a limitation as the \textbf{Probabilistic-Forecasting Gap}. Therefore, this leads to a critical question: \textbf{“How can we transform the deterministically reconstructed image into meaningful probabilistic forecasts that accurately reflect the uncertainty in future time series?”}

\par To bridge this gap, we introduce the ``\textbf{Multi-quantile Forecasting}'' approach for the time series converter, which extends the vision model's native capabilities (See bottom right part of Figure \ref{fig:main_framework}). Instead of modeling distributions explicitly (\eg, via parametric assumptions like Gaussian or Student's t \citep{DeepAR, Moirai} or via complex diffusion processes \citep{diffusion_TSF_1, diffusion_TSF_2}), we approximate the full forecast distribution through multiple quantile estimates, each reconstructed as a separate image.

\par Specifically, we equip the vision model with $h$ parallel heads, each tasked with reconstructing the masked image region corresponding to a target quantile level $\tau_k = \nicefrac{k}{h+1}$ for $k=1,\dots,h$. Each head is trained with the quantile loss (See Section~\ref{subsec:training_objective}), enabling specialization across the distribution—covering tails and central regions alike.

\par During the image-to-time-series decoding, each reconstructed image is split vertically into $M$ subfigures. Values from the designated RGB channel are extracted, resampled from $(\lfloor \nicefrac{W}{M} \rfloor, \nicefrac{W}{2})$ to $(P, \lfloor \nicefrac{T}{P} \rfloor)$, and reassembled into a $(T, M)$-shaped time series. This yields $h$ quantile forecasts, forming a complete probabilistic output.

\par Notably, our approach offers several advantages: (1) It enables \textit{seamless transfer learning}, repurposing pre-trained vision models for quantile forecasting with minimal architectural changes; (2) It performs \textit{distribution-free uncertainty modeling}, avoiding restrictive assumptions about output distributions; (3) It supports \textit{flexible quantile resolution}, allowing uncertainty granularity to be adjusted via the number of heads. Finally, the resulting framework thus unifies probabilistic and point forecasting: median quantiles (\eg, $\tau=0.5$) serve as robust point estimates, while the full set provides calibrated uncertainty intervals—making it adaptable to diverse downstream needs.

\subsection{Training Objective} \label{subsec:training_objective}

\par We train \model using a multi-quantile loss that jointly optimizes all $h$ forecasting heads. This objective supports probabilistic forecasting by supervising predictions across the full target distribution.

\par Specifically, let the target quantiles be $q_i = \frac{i}{h+1}$ for $i = 1, \dots, h$, with corresponding forecasts $\mX_{t:t+T}^{(i)}$ and ground truth $\hat{\mX}_{t:t+T}$. The quantile loss (or pinball loss) for head $i$ is defined as:
\begin{align*}
    \mathcal{L}_q = \frac{1}{h} \sum_{i=1}^{h} \max\left(q_i \cdot \mE_i,\ (q_i - 1) \cdot \mE_i\right),
    \quad \text{where} \quad \mE_i = \hat{\mX}_{t:t+T} - \mX_{t:t+T}^{(i)}.
\end{align*}

\par This loss ensures balanced optimization across quantiles, encouraging each head to specialize in its assigned level while sharing gradient signals across the ensemble. By avoiding point-estimate bias and making no distributional assumptions, it aligns naturally with our vision-based probabilistic framework and enables end-to-end training of the entire pipeline.

\section{Experiments} \label{sec:experiments}

\subsection{Experimental Setup}

\par \textbf{Training Dataset.}
We conduct continual pre-training of \model on the Large-scale Open Time Series Archive (LOTSA) \citep{Moirai, Moirai-MoE}, which is a diverse and multi-domain dataset containing over 231 billion observations. This scale and breadth can support robust temporal representation learning.

\par \textbf{Model Architecture.}
We train two variants of \model of different scales ($\model_{base}$ and $\model_{large}$), based on the 112M and 330M parameter \mae (base) and \mae (large) architectures \citep{MAE}. Both are initialized from ImageNet pre-trained weights.
Meanwhile, we set $h=9$ quantile heads targeting levels $\{10\%, 20\%, \dots, 90\%\}$ for probabilistic forecasting, balancing distributional coverage and model complexity.

\par \textbf{Training Process.}
Continual pre-training runs for 100,000 steps with a batch size of 512. We use the AdamW optimizer \citep{AdamW} (learning rate: $1\mathrm{e}{-4}$, weight decay: $1\mathrm{e}{-2}$, momentum terms: $\beta_1=0.9$, $\beta_2=0.98$), with a learning rate schedule combining 10,000-step linear warm-up and subsequent cosine annealing. All model parameters are fine-tuned to fully adapt visual representations to TSF.

\par \textbf{Evaluation Protocol.}
We follow recent TSFM research \citep{Moirai, VisionTS, Moirai-MoE} and evaluate on three established benchmarks: Monash \citep{Monash}, Long-term Time Series Forecasting (LTSF) \citep{Autoformer}, and Probabilistic Forecasting (PF) \citep{Moirai}, all compatible with LOTSA to avoid data leakage. We compare \model against state-of-the-art foundation models, deep learning, and classical baselines (details in Appendix~\ref{sec:app_benchmark_baseline}). Notably, our key comparisons include: (1) \visionts (ImageNet-pretrained) — to assess the impact of CPT on temporal data adaptation; and (2) \moirai (LOTSA-pretrained) — to evaluate the benefit of visual pre-training. This dual comparison isolates the roles of modality transfer and temporal scaling in foundation models.

\subsection{In-distribution Forecasting}

\paragraph{Monash Time Series Forecasting.}
\par We evaluate in-distribution performance on a total of 29 datasets from the Monash benchmark \citep{Monash} (details in Appendix~\ref{subsec:app_benchmarks}). 
To ensure a fair and rigorous comparison, the pre-training dataset LOTSA includes only the training portions of these series, with test sets held out for evaluation.

\par
\begin{figure*}[t]
    \centering  %
    \includegraphics[width=0.9\textwidth]{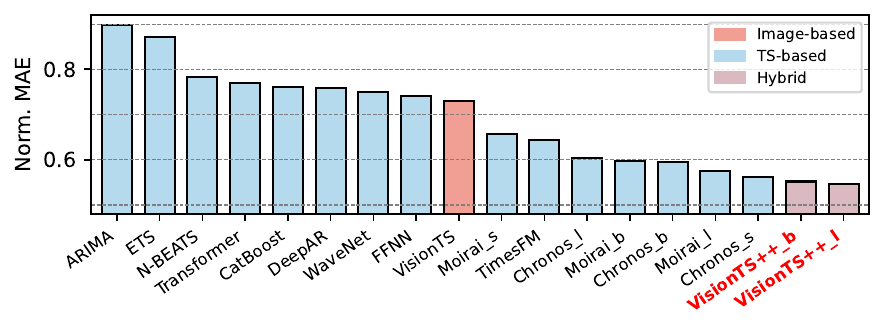}
    \vspace{-0.5em}
    \caption{Normalized MAE results on Monash Benchmark, with full results in Table \ref{tab:monash_full} (Appendix \ref{subsec:app_monash_full}). Model sizes are denoted as: \textbf{s} (small), \textbf{b} (base), \textbf{l} (large).}
    \vspace{-0.5em}
    \label{fig:monash_results}
\end{figure*}

 Figure~\ref{fig:monash_results} reports the normalized mean absolute error (nMAE), defined as the geometric mean of each model's MAE scaled by the naive forecasting baseline per dataset. 
The results show that \model achieves state-of-the-art performance across all models. It outperforms both dataset-specific models and the original \visionts by over 23.2\%, validating the effectiveness of our conversion and pre-training paradigm. 
Notably, \model also surpasses \moirai—a foundation model trained on the same data—across all three sizes. This improvement, under identical training data and evaluation conditions, indicates that \model's ImageNet-pretrained visual knowledge provides a more effective initialization than training from scratch. 
The transferred visual representations enhance feature learning efficiency and in-distribution forecasting, demonstrating the value of cross-modal pre-training.

\begin{table}[htbp]
  \centering
  \caption{Zero-shot results on LTSF benchmark of base and large models, averaged over four prediction lengths \{96, 192, 336, 720\}. Full results are in \cref{tab:ltsf_full} (\cref{subsec:app_ltsf_full}). Time-MoE, Timer, and TimesFM are excluded in Electricity and Weather since time series were used in their pre-training.}
  \vspace{-0.2em}
  \resizebox{\linewidth}{!}{
    \begin{tabular}{ccccccccccccccccccccccccc}
    \toprule
    & \textbf{Pre-train} & \multicolumn{3}{c}{\emoji{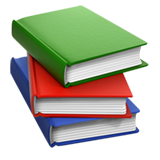} \textbf{Hybrid}} &       & \multicolumn{1}{c}{\emoji{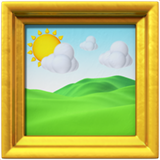} \textbf{Images}} &       & \multicolumn{17}{c}{\emoji{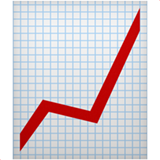} \textbf{Time-Series}} \\
\cmidrule{3-5}\cmidrule{7-7}\cmidrule{9-25}    \textbf{Dataset} & \textbf{Method} & \textcolor{red}{$\textbf{VisionTS++}_{l}$} &       & \textcolor{red}{$\textbf{VisionTS++}_{b}$} &       & \textbf{VisionTS} &       & $\textbf{Time-MoE}_{s}$ &       & $\textbf{Time-MoE}_{b}$ &       & $\textbf{Chronos}_{s}$ &       & $\textbf{Chronos}_{l}$ &       & $\textbf{Moirai}_{s}$ &       & $\textbf{Moirai}_{l}$ &       & \textbf{Moment} &       & $\textbf{Timer}_{28B}$ &       & \textbf{TimesFM} \\
    \midrule
    \multirow{2}[2]{*}{ETTm1} & MSE   & \textbf{0.354}  &       & 0.360  &       & 0.374  &       & 0.394  &       & 0.376  &       & 0.640  &       & 0.556  &       & 0.448  &       & 0.390  &       & 0.670  &       & 0.487  &       & 0.433  \\
          & MAE   & \textbf{0.369}  &       & 0.372  &       & 0.372  &       & 0.416  &       & 0.406  &       & 0.500  &       & 0.465  &       & 0.410  &       & 0.389  &       & 0.537  &       & 0.457  &       & 0.419  \\
    \midrule
    \multirow{2}[2]{*}{ETTm2} & MSE   & \textbf{0.244 } &       & \textbf{0.244 } &       & 0.282  &       & 0.318  &       & 0.316  &       & 0.349  &       & 0.295  &       & 0.300  &       & 0.276  &       & 0.317  &       & 0.316  &       & 0.328  \\
          & MAE   & \textbf{0.298 } &       & \textbf{0.298 } &       & 0.321  &       & 0.366  &       & 0.361  &       & 0.380  &       & 0.338  &       & 0.341  &       & 0.320  &       & 0.366  &       & 0.371  &       & 0.347  \\
    \midrule
    \multirow{2}[2]{*}{ETTh1} & MSE   & 0.403  &       & 0.402  &       & \textbf{0.390 } &       & 0.400  &       & 0.394  &       & 0.545  &       & 0.589  &       & 0.400  &       & 0.510  &       & 0.684  &       & 0.444  &       & 0.473  \\
          & MAE   & 0.418  &       & 0.416  &       & \textbf{0.414 } &       & 0.424  &       & 0.420  &       & 0.472  &       & 0.466  &       & 0.424  &       & 0.469  &       & 0.566  &       & 0.457  &       & 0.444  \\
    \midrule
    \multirow{2}[2]{*}{ETTh2} & MSE   & \textbf{0.327}  &       & 0.333  &       & 0.333  &       & 0.367  &       & 0.405  &       & 0.424  &       & 0.455  &       & 0.341  &       & 0.354  &       & 0.362  &       & 0.358  &       & 0.392  \\
          & MAE   & \textbf{0.365}  &       & 0.370  &       & 0.375  &       & 0.404  &       & 0.415  &       & 0.430  &       & 0.427  &       & 0.379  &       & 0.377  &       & 0.410  &       & 0.407  &       & 0.406  \\
    \midrule
    \multirow{2}[2]{*}{Electricity} & MSE   & \textbf{0.181}  &       & 0.184  &       & 0.207  &       & - &  & - &  & 0.220  &       & 0.204  &       & 0.233  &       & 0.188  &       & 0.765  &       & - &  & - \\
          & MAE   & \textbf{0.264}  &       & 0.265  &       & 0.294  &       & - &  & - &  & 0.284  &       & 0.274  &       & 0.320  &       & 0.273  &       & 0.687  &       & - &  & - \\
    \midrule
    \multirow{2}[1]{*}{Weather} & MSE   & 0.226  &       & \textbf{0.222 } &       & 0.269  &       & 0.266  &       & 0.270  &       & 0.300  &       & 0.279  &       & 0.242  &       & 0.260  &       & 0.294  &       & 0.304  &       & - \\
          & MAE   & 0.243  &       & \textbf{0.241 } &       & 0.292  &       & 0.297  &       & 0.300  &       & 0.318  &       & 0.306  &       & 0.267  &       & 0.275  &       & 0.326  &       & 0.331  &       & - \\
    \midrule
    \multirow{2}[0]{*}{\textbf{Average}} & MSE   & \textbf{0.289}  &       & 0.291  &       & 0.309  &       & - &  & - &  & 0.413  &       & 0.396  &       & 0.327  &       & 0.329  &       & 0.515  &       & - &  & - \\
          & MAE   & \textbf{0.326}  &       & 0.327  &       & 0.345  &       & - &  & - &  & 0.397  &       & 0.379  &       & 0.357  &       & 0.350  &       & 0.482  &       & - &  & - \\
    \rowc
    \multicolumn{2}{c}{$\bf 1^{st}$ \textbf{count}} & \textcolor{red}{\textbf{10}}    &       & \textcolor{red}{\textbf{4}}     &       & 2     &       & 0     &       & 0     &       & 0     &       & 0     &       & 0     &       & 0     &       & 0     &       & 0     &       & 0 \\
    \bottomrule
    \end{tabular}
  }
  \label{tab:ltsf_main_text}%
  \vspace{-0.5em}
\end{table}%

\subsection{Out-of-distribution Forecasting}

\par To further evaluate the generalization capability, we conduct out-of-distribution forecasting (\ie, zero-shot forecasting) experiments on two benchmarks—Long-term Time Series Forecasting (LTSF) \citep{Autoformer} and Probabilistic Forecasting (PF) \citep{Moirai}—where neither training nor test data overlap with the pre-training corpus LOTSA. This setup assesses the model's ability to transfer learned representations to unseen domains.

\paragraph{Long-term Time Series Forecasting (LTSF).}
\par We compare \model against state-of-the-art TSFMs including \visionts \citep{VisionTS}, Time-MoE \citep{Time-MoE}, Moirai \citep{Moirai}, Chronos \citep{Chronos}, \etc
Table~\ref{tab:ltsf_main_text} reports averaged Mean Squared Error (MSE) and Mean Absolute Error (MAE) across four prediction lengths $\{96, 192, 336, 720\}$ (full results in Table~\ref{tab:ltsf_full} in Appendix~\ref{subsec:app_ltsf_full}).

\par The results show that \model achieves the best performance in 12 out of 14 settings. It improves over \visionts by 6\% in average MSE, confirming that our image conversion and continual pre-training preserve visual priors while enhancing temporal modeling. 
Notably, \model outperforms specialized TSFMs by 6\%–44\% in MSE, demonstrating that with appropriate adaptation, vision-based models can surpass domain-specific architectures in long-term forecasting.

\begin{table}[htpb]
  \centering
  \caption{Zero-shot results on the probabilistic forecasting benchmark. Best results are in \textbf{bold}.} %
  \vspace{-0.2em}
  \resizebox{\columnwidth}{!}{
    \begin{tabular}{ccccccccccccccc}
    \toprule
    &  & \multicolumn{6}{c}{\textbf{Zero-shot}} & \multicolumn{4}{c}{\textbf{Full-shot}} & \multicolumn{2}{c}{\textbf{Baseline}} \\
    \cmidrule(lr){3-8} \cmidrule(lr){9-12} \cmidrule(lr){13-14}
          \textbf{Dataset} & \textbf{Method} & \textcolor{red}{$\textbf{VisionTS++}_{l}$} & \textcolor{red}{$\textbf{VisionTS++}_{b}$} & \textbf{VisionTS}  & $\textbf{Moirai}_{s}$ & $\textbf{Moirai}_{b}$ & $\textbf{Moirai}_{l}$ & \textbf{PatchTST} & \textbf{TiDE} & \textbf{TFT} & \textbf{DeepAR} & \textbf{AutoARIMA} & \textbf{Seasonal Naive} \\
    \midrule
    \multirow{2}[2]{*}{Electricity} & \textbf{CRPS} & \textbf{0.041} & 0.042 & 0.068 & 0.072 & 0.055 & 0.050 & 0.052{\scriptsize$\pm$}0.00 & 0.048{\scriptsize$\pm$}0.00 & 0.050{\scriptsize$\pm$}0.00 & 0.065{\scriptsize$\pm$}0.01 & 0.327 & 0.070 \\
          & \textbf{MASE} & 0.635 & \textbf{0.631} & 0.755 & 0.981 & 0.792 & 0.751 & 0.753{\scriptsize$\pm$}0.01 & 0.706{\scriptsize$\pm$}0.02 & 0.747{\scriptsize$\pm$}0.03 & 0.844{\scriptsize$\pm$}0.16 & 3.229 & 0.881 \\
    \midrule
    \multirow{2}[2]{*}{Solar} & \textbf{CRPS} & \textbf{0.353} & \textbf{0.353} & 0.502 & 0.471 & 0.419 & 0.406 & 0.518{\scriptsize$\pm$}0.09 & 0.420{\scriptsize$\pm$}0.00 & 0.446{\scriptsize$\pm$}0.03 & 0.431{\scriptsize$\pm$}0.01 & 1.055 & 0.512 \\
          & \textbf{MASE} & \textbf{1.135} & 1.155 & 1.141 & 1.465 & 1.292 & 1.237 & 1.607{\scriptsize$\pm$}0.25 & 1.265{\scriptsize$\pm$}0.02 & 1.399{\scriptsize$\pm$}0.11 & 1.222{\scriptsize$\pm$}0.01 & 2.583 & 1.203 \\
    \midrule
    \multirow{2}[2]{*}{Walmart} & \textbf{CRPS} & \textbf{0.061} & 0.064 & 0.121 & 0.103 & 0.093 & 0.098 & 0.082{\scriptsize$\pm$}0.01 & 0.077{\scriptsize$\pm$}0.00 & 0.087{\scriptsize$\pm$}0.00 & 0.121{\scriptsize$\pm$}0.00 & 0.124 & 0.151 \\
          & \textbf{MASE} & \textbf{0.684} & 0.689 & 0.949 & 1.048 & 0.964 & 1.007 & 0.867{\scriptsize$\pm$}0.09 & 0.814{\scriptsize$\pm$}0.01 & 0.948{\scriptsize$\pm$}0.02 & 1.193{\scriptsize$\pm$}0.02 & 1.131 & 1.236 \\
    \midrule
    \multirow{2}[2]{*}{Weather} & \textbf{CRPS} & \textbf{0.038} & \textbf{0.038} & 0.056 & 0.049 & 0.041 & 0.051 & 0.059{\scriptsize$\pm$}0.01 & 0.054{\scriptsize$\pm$}0.00 & 0.043{\scriptsize$\pm$}0.00 & 0.132{\scriptsize$\pm$}0.11 & 0.252 & 0.068 \\
          & \textbf{MASE} & 0.449 & \textbf{0.447} & 0.737 & 0.521 & 0.487 & 0.515 & 0.844{\scriptsize$\pm$}0.19 & 0.832{\scriptsize$\pm$}0.13 & 0.692{\scriptsize$\pm$}0.02 & 3.170{\scriptsize$\pm$}3.47 & 0.938 & 0.782 \\
    \midrule
    \multirow{2}[2]{*}{Istanbul Traffic} & \textbf{CRPS} & \textbf{0.105} & 0.115 & 0.198 & 0.173 & 0.116 & 0.112 & 0.112{\scriptsize$\pm$}0.00 & 0.110{\scriptsize$\pm$}0.01 & 0.110{\scriptsize$\pm$}0.01 & 0.108{\scriptsize$\pm$}0.00 & 0.589 & 0.257 \\
          & \textbf{MASE} & \textbf{0.590} & 0.616 & 0.706 & 0.990 & 0.644 & 0.631 & 0.653{\scriptsize$\pm$}0.02 & 0.618{\scriptsize$\pm$}0.03 & 0.620{\scriptsize$\pm$}0.03 & 0.613{\scriptsize$\pm$}0.03 & 3.358 & 1.137 \\
    \midrule
    \multirow{2}[2]{*}{Turkey Power} & \textbf{CRPS} & 0.038 & \textbf{0.036} & 0.052 & 0.048 & 0.040 & \textbf{0.036} & 0.054{\scriptsize$\pm$}0.01 & 0.046{\scriptsize$\pm$}0.01 & 0.039{\scriptsize$\pm$}0.00 & 0.066{\scriptsize$\pm$}0.02 & 0.116 & 0.085 \\
          & \textbf{MASE} & 0.752 & \textbf{0.737} & 0.856 & 0.948 & 0.888 & 0.870 & 1.234{\scriptsize$\pm$}0.12 & 0.904{\scriptsize$\pm$}0.02 & 0.890{\scriptsize$\pm$}0.05 & 1.395{\scriptsize$\pm$}0.30 & 1.700 & 0.906 \\
    \midrule
    \multirow{2}[2]{*}{\textbf{Norm.}} & CRPS & \textbf{0.506} & 0.515 & 0.816 & 0.749 & 0.608 & 0.609 & 0.679 & 0.612 & 0.595 & 0.857 & 2.123 & 1.000 \\
          & MASE & \textbf{0.673} & 0.677 & 0.838 & 0.942 & 0.799 & 0.794 & 0.937 & 0.827 & 0.843 & 1.211 & 1.906 & 1.000 \\
    \rowc
    \multicolumn{2}{c}{\textbf{$\bf 1^{st}$ count}} & \textcolor{red}{\textbf{10}} & \textcolor{red}{\textbf{6}} & 0  & 0  & 0  & 1  & 0  & 0  & 0  & 0  & 0  & 0 \\
    \bottomrule
    \end{tabular}%
  }
  \label{tab:pf_main_text}%
  \vspace{-0.5em}
\end{table}%

\paragraph{Probabilistic Forecasting (PF).}
\par We further evaluate probabilistic forecasting on six real-world datasets (across energy, transport, climate, and sales domains) using the Continuous Ranked Probability Score (CRPS), along with MASE for point forecasting.

\par Based on results in Table~\ref{tab:pf_main_text}, \model ranks first in all scenarios across both metrics. It significantly improves upon \visionts, validating the effectiveness of the multi-quantile forecasting design. More importantly, \model outperforms not only zero-shot but also full-shot baselines—despite receiving no dataset-specific training—highlighting its strong generalization. These results indicate that, with appropriate continual pre-training, vision-based TSFM can achieve SOTA zero-shot performance in probabilistic forecasting.

\paragraph{GIFT-Eval Benchmark.}
\par Additionally, we evaluate on the General Time Series Forecasting Model Evaluation (GIFT-Eval) benchmark \citep{gift-eval}, which comprises 23 datasets across 7 domains. To ensure consistent evaluation, we re-train a version of \model using their ``GiftEvalPretrain'' dataset. strictly avoiding potential data leakage. We compare our model against TSFMs that similarly avoid data leakage, with baseline models cut-off as of the submission of \model.

\par
\begin{figure*}[t]
    \centering  %
    \includegraphics[width=0.98\textwidth]{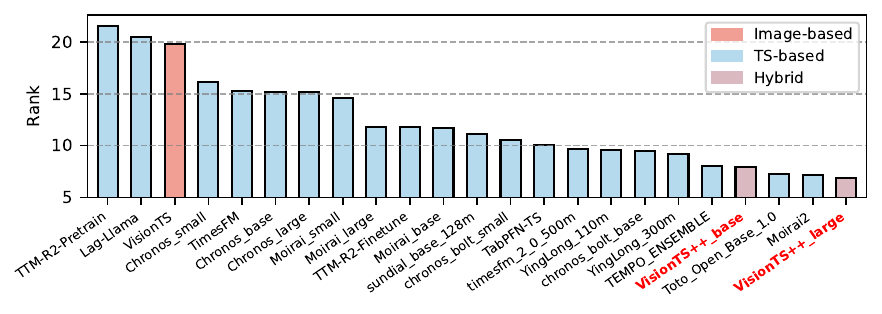}
    \vspace{-0.5em}
    \caption{Ranks on GIFT-Eval Benchmark (cut-off at 2025/08).}
    \vspace{-0.5em}
    \label{fig:gift_eval_results}
\end{figure*}

 Based on results in Figure \ref{fig:gift_eval_results}, \model-large achieves the top rank under the aggregated ranking combining CRPS and MASE metrics, with the base model also ranking highly. Since GIFT-Eval includes both univariate and multivariate, as well as deterministic and probabilistic forecasting, this result demonstrates that \model effectively generalizes across diverse domains and supports a wide range of forecasting scenarios.

\subsection{Further Analysis on \model} \label{subsec:further_analysis}

\paragraph{Random Initialization.}
\par To assess the importance of visual knowledge in the \mae backbone, we compare \model using ImageNet-pretrained weights versus random parameters for initialization before conducting continual pre-training. The results are reported in Table~\ref{tab:ablation_rand_init} in Appedix \ref{subsec:app_ablation_full} due to space limit. 

\par These results reveal that the randomly initialized variant suffers a nearly 30\% degradation in aggregated performance. This significant drop confirms that original visual representations provide an essential inductive bias for TSF, and that our continual pre-training effectively adapts—rather than overwrites—these features for time series.

\paragraph{Ablation Study.}
\par We further ablate key components of \model (presented in Table~\ref{tab:ablation_all}) in Appedix \ref{subsec:app_ablation_full}, demonstrating the contribution of each design:

\begin{itemize}[leftmargin=*]
    \item \textbf{Vision-model-based Filtering.}
    Removing this module leads to a 7\% performance drop. It mitigates modality mismatch by filtering out extreme values that distort pixel-aligned visual representations, ensuring compatibility with the pre-trained backbone.

    \item \textbf{Colorized Multivariate Conversion.} 
    Replacing RGB-encoded multivariate subfigures with grayscale univariate inputs (as in \visionts) increases MSE by 12\%. The colorization strategy leverages the vision model’s sensitivity to spatial and chromatic structure, enhancing cross-variate dependency modeling through multi-object analysis.
    
    \item \textbf{Multi-quantile Forecasting.} 
    Collapsing to a single forecasting head degrades the probabilistic performance by over 10\%. This validates that our unified design, which constructs and repurposes multiple \mae’s pixel reconstruction heads for quantile estimation, enables effective distributional forecasting.

\end{itemize}

\section{Conclusions and Future Work} \label{conclusion}

In this paper, we propose \model, a time series foundation model based on the continual pre-training of a vision foundation model on large-scale time series data. 
To bridge critical inherent gaps between images and time series, we introduce three key components, including vision-model-based filtering, colorized multivariate conversion, and multi-quantile forecasting. 
These designs enable effective adaptation of visual representations to time series patterns without modifying the underlying model architecture.

Extensive evaluation shows that \model achieves state-of-the-art performance across both in-distribution (Monash) and out-of-distribution (LTSF, PF) benchmarks, outperforming specialized TSFMs. 
These results demonstrate that pre-trained visual representations, when appropriately aligned with time series data, can serve as a powerful foundation for forecasting. 
Notably, our approach preserves valuable cross-modal knowledge while enabling robust temporal generalization—highlighting the potential of vision-based models in time series understanding.

Future work includes exploring larger-scale multi-modal pre-training, extending the framework to other time series tasks such as classification and anomaly detection, and investigating dynamic filtering mechanisms for diverse data regimes.
Additionally, further integration with video foundation models may exploit spatio-temporal structure, advancing in more powerful universal models capable of unified visual and temporal understanding.

\bibliographystyle{iclr2026_conference}
\bibliography{iclr2026_conference}

\appendix

\numberwithin{equation}{section}
\section*{Appendix}

\section{Related Works} \label{related-works}

\subsection{Time Series Foundation Models}
\par Recent advances in time series forecasting have seen the emergence of time series foundation models (TSFMs) as powerful zero-shot forecasting tools. Unlike traditional dataset-specific models (\eg, PatchTST \citep{PatchTST}, TiDE \citep{TiDE}, FEDformer \citep{FEDformer}) that require training on target datasets, TSFMs leverage large-scale pre-training to achieve cross-domain generalization. 
These models are typically pre-trained on diverse real-world time series datasets across diverse domains \citep{Moment, Timer, TimesFM, TimeSiam, GTT} or pre-trained on synthetic time series data \citep{liu2025syntheticsurvey, SyntheticTS, ViTime}.
Notable examples include Moirai \citep{Moirai}, which assembles a data archive of 231 billion observations across nine domains to train encoder-based models of varying scales, demonstrating strong zero-shot capabilities. 
Other foundation models with mostly encoder-based or decoder-based architectures have shown similar success, including Chronos \citep{Chronos}, TimesFM \citep{TimesFM}, Timer \citep{Timer}, Moment \citep{Moment}, and Time-MoE \citep{Time-MoE}.
However, developing an effective TSFM faces significant challenges due to the inherent heterogeneity and high noise in time series data, thus demanding the construction of high-quality training datasets.

\subsection{Vision Models for Time Series Analysis}

\par The exploration of vision-model-based approaches for time series analysis has significantly progressed in recent years. 
Early works demonstrate that encoding time series as images enables effective application of convolutional neural networks (CNNs) for both classification \citep{wang2015imaging, wang2015spatially, hatami2018classification} and forecasting tasks \citep{li2020forecasting, sood2021visual, semenoglou2023image}. 
More recent advances have started to leverage pre-trained visual foundation models or vision-language models for time series analysis. 
For instance, AST \citet{AST} adopts DeiT \citep{DeiT} for time series classification, and ViTST \citep{ViTST} utilizes pre-trained vision transformers (ViTs) \citep{ViT} and swin transformers \citep{Swin_transformer} to further explore this direction.
Other works, such as \citet{wimmer2023leveraging} and \citet{zhang2023insight}, explore the use of vision-language models for feature extraction and textual description generation.
Moreover, ViTime \citep{ViTime} generates synthetic time series data and converts them into line plots for pre-training vision models such as ViT.
ImagenTime \citep{ImagenTime} introduces a unified generative framework by transforming time series into images via invertible methods like delay embedding and STFT, enabling them to leverage advanced vision diffusion models for generation, interpolation, and extrapolation tasks.
Several recent surveys \citep{vision4ts_survey, vision4ts_survey_2, vision4ts_survey_3, vision4ts_survey_4} have also discussed the application of vision models or multi-modal approaches in time series analysis.
For example, Vision4TS \citep{vision4ts_survey} summarizes crucial techniques including time-series-to-image transformation, image pre-processing, and modeling strategies for imaged time series.
The most relevant approach is \visionts \citep{VisionTS}, which reformulates time series forecasting as a patch-level image reconstruction task and leverages the visual \mae model as the backbone.

\par However, although these methods establish preliminary connections between visual and time series domains, they fail to sufficiently address some critical modality gaps. 
To the best of our knowledge, we are the first to propose a competitive TSFM through continual pretraining on vision backbones, thus better enhancing the transferability between two modalities.

\section{Benchmarks \& Baselines} \label{sec:app_benchmark_baseline}

\subsection{Benchmarks} \label{subsec:app_benchmarks}

\paragraph{Monash Benchmark} 
\par Following \citet{Moirai}, we tested 29 Monash datasets \citep{Monash} using GluonTS \citep{GluonTS}, including M1 Monthly, M3 Monthly, M3 Other, M4 Monthly, M4 Weekly, M4 Daily, M4 Hourly, Tourism Quarterly, Tourism Monthly, CIF 2016, Australian Electricity Demand, Bitcoin, Pedestrian Counts, Vehicle Trips, KDD Cup, Weather, NN5 Daily, NN5 Weekly, Carparts, FRED-MD, Traffic Hourly, Traffic Weekly, Rideshare, Hospital, COVID Deaths, Temperature Rain, Sunspot, Saugeen River Flow, and US Births. Performance is assessed using Mean Absolute Error (MAE) metric.

\paragraph{Probabilistic Forecasting Benchmark} The Probabilistic Forecasting (PF) Benchmark \citep{Moirai} consists of 6 datasets across energy, transport, climate, and sales domains, including Electricity, Solar, Walmart, Weather, Istanbul Traffic, and Turkey Power.
Performance is assessed using Continuous Ranked Probability Score (CRPS) and Mean Absolute Scaled Error (MASE) metrics.

\paragraph{Long-Term TSF Benchmark} We evaluate our model on 6 widely used long-term TSF datasets \citep{Informer, Autoformer}, including ETTh1, ETTh2, ETTm1, ETTm2, Electricity, and Weather. Performance is assessed using Mean Squared Error (MSE) and Mean Absolute Error (MAE) metrics. 

\paragraph{GIFT-Eval Benchmark} \citet{gift-eval} introduces the General Time Series Forecasting Model Evaluation (GIFT-Eval), encompasses 23 datasets over 144,000 time series and 177 million data points, spanning 7 domains, 10 frequencies, multivariate inputs, and prediction lengths ranging from short to long-term forecasts.

\begin{table}[htp]
  \centering
  \caption{Full results of Monash Time Series Forecasting Benchmark. MAE is reported.}
  \resizebox{\columnwidth}{!}{
    \begin{tabular}{lccccccccccccccccccccc}
    \toprule
          & \textcolor{red}{$\textbf{VisionTS++}_{l}$} & \textcolor{red}{$\textbf{VisionTS++}_{b}$} & \textbf{VisionTS (z.s.)} & \textbf{LLMTime (z.s.)} & $\textbf{Moirai}_{s}$ & $\textbf{Moirai}_{b}$ & $\textbf{Moirai}_{l}$ & \textbf{Naive} & \textbf{SES} & \textbf{Theta} & \textbf{TBATS} & \textbf{ETS} & \textbf{(DHR-)ARIMA} & \textbf{PR} & \textbf{CatBoost} & \textbf{FFNN} & \textbf{DeepAR} & \textbf{N-BEATS} & \textbf{WaveNet} & \textbf{Transformer} \\
    \midrule
    M1 Monthly & 1919.97 & 1,846.05 & 1987.69 & 2562.84 & 2,082.26 & 2,068.63 & 1,983.18 & 2,707.75 & 2,259.04 & 2,166.18 & 2,237.50 & 1,905.28 & 2,080.13 & 2,088.25 & 2,052.32 & 2,162.58 & 1,860.81 & 1,820.37 & 2,184.42 & 2,723.88 \\
    M3 Monthly & 591.44 & 581.68 & 737.93 & 877.97 & 713.41 & 658.17 & 664.03 & 837.14 & 743.41 & 623.71 & 630.59 & 626.46 & 654.8 & 692.97 & 732   & 692.48 & 728.81 & 648.6 & 699.3 & 798.38 \\
    M3 Other & 180.99 & 186.13 & 315.85 & 300.3 & 263.54 & 198.62 & 202.41 & 278.43 & 277.83 & 215.35 & 189.42 & 194.98 & 193.02 & 234.43 & 318.13 & 240.17 & 247.56 & 221.85 & 245.29 & 239.24 \\
    M4 Monthly & 533.16 & 533.74 & 666.54 & 728.27 & 597.6 & 592.09 & 584.36 & 671.27 & 625.24 & 563.58 & 589.52 & 582.6 & 575.36 & 596.19 & 611.69 & 612.52 & 615.22 & 578.48 & 655.51 & 780.47 \\
    M4 Weekly & 281.76 & 280.88 & 404.23 & 518.44 & 339.76 & 328.08 & 301.52 & 347.99 & 336.82 & 333.32 & 296.15 & 335.66 & 321.61 & 293.21 & 364.65 & 338.37 & 351.78 & 277.73 & 359.46 & 378.89 \\
    M4 Daily & 190.54 & 172.31 & 215.63 & 266.52 & 189.1 & 192.66 & 189.78 & 180.83 & 178.27 & 178.86 & 176.6 & 193.26 & 179.67 & 181.92 & 231.36 & 177.91 & 299.79 & 190.44 & 189.47 & 201.08 \\
    M4 Hourly & 169.17 & 202.99 & 288.37 & 576.06 & 268.04 & 209.87 & 197.79 & 1,218.06 & 1,218.06 & 1,220.97 & 386.27 & 3,358.10 & 1,310.85 & 257.39 & 285.35 & 385.49 & 886.02 & 425.75 & 393.63 & 320.54 \\
    Tourism Quarterly & 5823.41 & 6,055.50 & 12931.88 & 16918.86 & 18,352.44 & 17,196.86 & 15,820.02 & 15,845.10 & 15,014.19 & 7,656.49 & 9,972.42 & 8,925.52 & 10,475.47 & 9,092.58 & 10,267.97 & 8,981.04 & 9,511.37 & 8,640.56 & 9,137.12 & 9,521.67 \\
    Tourism Monthly & 1667.65 & 2,065.71 & 2560.19 & 5608.61 & 3,569.85 & 2,862.06 & 2,688.55 & 5,636.83 & 5,302.10 & 2,069.96 & 2,940.08 & 2,004.51 & 2,536.77 & 2,187.28 & 2,537.04 & 2,022.21 & 1,871.69 & 2,003.02 & 2,095.13 & 2,146.98 \\
    CIF 2016 & 5664485.37 & 549,318.73 & 570907.24  & 599313.8 & 655,888.58 & 539,222.03 & 695,156.92 & 578,596.53 & 581,875.97 & 714,818.58 & 855,578.40 & 642,421.42 & 469,059.49 & 563,205.57 & 603,551.30 & 1,495,923.44 & 3,200,418.00 & 679,034.80 & 5,998,224.62 & 4,057,973.00 \\
    Aus. Elec. Demand & 180.99 & 226.31 & 237.44 & 760.81 & 266.57 & 201.39 & 177.68 & 659.6 & 659.6 & 665.04 & 370.74 & 1,282.99 & 1,045.92 & 247.18 & 241.77 & 258.76 & 302.41 & 213.83 & 227.5 & 231.45 \\
    Bitcoin & 1.85E+18 & 1.81E+18 & 2.33E+18 & 1.74E+18 & 1.76E+18 & 1.62E+18 & 1.87E+18 & 7.78E+17 & 5.33E+18 & 5.33E+18 & 9.90E+17 & 1.10E+18 & 3.62E+18 & 6.66E+17 & 1.93E+18 & 1.45E+18 & 1.95E+18 & 1.06E+18 & 2.46E+18 & 2.61E+18 \\
    Pedestrian Counts & 61.47 & 62.55 & 52.01 & 97.77 & 54.88 & 54.08 & 41.66 & 170.88 & 170.87 & 170.94 & 222.38 & 216.5 & 635.16 & 44.18 & 43.41 & 46.41 & 44.78 & 66.84 & 46.46 & 47.29 \\
    Vehicle Trips & 20.67 & 19.98 & 22.08 & 31.48 & 24.46 & 23.17 & 21.85 & 31.42 & 29.98 & 30.76 & 21.21 & 30.95 & 30.07 & 27.24 & 22.61 & 22.93 & 22    & 28.16 & 24.15 & 28.01 \\
    KDD cup & 38.75 & 38.89 & 38.16 & 42.72 & 39.81 & 38.66 & 39.09 & 42.13 & 42.04 & 42.06 & 39.2  & 44.88 & 52.2  & 36.85 & 34.82 & 37.16 & 48.98 & 49.1  & 37.08 & 44.46 \\
    Weather & 1.73 & 1.73 & 2.06 & 2.17 & 1.96  & 1.8   & 1.75  & 2.36  & 2.24  & 2.51  & 2.3   & 2.35  & 2.45  & 8.17  & 2.51  & 2.09  & 2.02  & 2.34  & 2.29  & 2.03  \\
    NN5 Daily & 3.51 & 3.41 & 3.51 & 7.1 & 5.37  & 4.26  & 3.77  & 8.26  & 6.63  & 3.8   & 3.7   & 3.72  & 4.41  & 5.47  & 4.22  & 4.06  & 3.94  & 4.92  & 3.97  & 4.16  \\
    NN5 Weekly & 14.84 & 14.12 & 14.67 & 15.76 & 15.07 & 16.42 & 15.3  & 16.71 & 15.66 & 15.3  & 14.98 & 15.7  & 15.38 & 14.94 & 15.29 & 15.02 & 14.69 & 14.19 & 19.34 & 20.34 \\
    Carparts & 0.44 & 0.43 & 0.58 & 0.44 & 0.53  & 0.47  & 0.49  & 0.65  & 0.55  & 0.53  & 0.58  & 0.56  & 0.56  & 0.41  & 0.53  & 0.39  & 0.39  & 0.98  & 0.4   & 0.39  \\
    FRED-MD & 2722.75 & 2,347.09 & 1893.67 & 2804.64 & 2,568.48 & 2,679.29 & 2,792.55 & 2,825.67 & 2,798.22 & 3,492.84 & 1,989.97 & 2,041.42 & 2,957.11 & 8,921.94 & 2,475.68 & 2,339.57 & 4,264.36 & 2,557.80 & 2,508.40 & 4,666.04 \\
    Traffic Hourly & 0.013 & 0.016 & 0.01 & 0.03 & 0.02  & 0.02  & 0.01  & 0.03  & 0.03  & 0.03  & 0.04  & 0.03  & 0.04  & 0.02  & 0.02  & 0.01  & 0.01  & 0.02  & 0.02  & 0.01  \\
    Traffic Weekly & 1.08 & 1.07 & 1.14 & 1.15 & 1.17  & 1.14  & 1.13  & 1.19  & 1.12  & 1.13  & 1.17  & 1.14  & 1.22  & 1.13  & 1.17  & 1.15  & 1.18  & 1.11  & 1.2   & 1.42  \\
    Rideshare & 1.37 & 1.36 & 5.92 & 6.28 & 1.35  & 1.39  & 1.29  & 6.29  & 6.29  & 7.62  & 6.45  & 6.29  & 3.37  & 6.3   & 6.07  & 6.59  & 6.28  & 5.55  & 2.75  & 6.29  \\
    Hospital & 17.30 & 17.00 & 19.36 & 25.68 & 23    & 19.4  & 19.44 & 24.07 & 21.76 & 18.54 & 17.43 & 17.97 & 19.6  & 19.24 & 19.17 & 22.86 & 18.25 & 20.18 & 19.35 & 36.19 \\
    COVID Deaths & 114.97 & 151.53 & 137.51 & 653.31 & 124.32 & 126.11 & 117.11 & 353.71 & 353.71 & 321.32 & 96.29 & 85.59 & 85.77 & 347.98 & 475.15 & 144.14 & 201.98 & 158.81 & 1,049.48 & 408.66 \\
    Temperature Rain & 4.83 & 5.17 & 6.37 & 6.37 & 5.3   & 5.08  & 5.27  & 9.39  & 8.18  & 8.22  & 7.14  & 8.21  & 7.19  & 6.13  & 6.76  & 5.56  & 5.37  & 7.28  & 5.81  & 5.24  \\
    Sunspot & 0.25 & 0.19 & 2.81 & 5.07 & 0.11  & 0.08  & 0.13  & 3.93  & 4.93  & 4.93  & 2.57  & 4.93  & 2.57  & 3.83  & 2.27  & 7.97  & 0.77  & 14.47 & 0.17  & 0.13  \\
    Saugeen River Flow & 23.24 & 24.24 & 30.22 & 34.84 & 24.07 & 24.4  & 24.76 & 21.5  & 21.5  & 21.49 & 22.26 & 30.69 & 22.38 & 25.24 & 21.28 & 22.98 & 23.51 & 27.92 & 22.17 & 28.06  \\
    US Births & 420.22 & 411.48 & 519.94 & 1374.99 & 872.51 & 624.3 & 476.5 & 1,152.67 & 1,192.20 & 586.93 & 399   & 419.73 & 526.33 & 574.93 & 441.7 & 557.87 & 424.93 & 422   & 504.4 & 452.87 \\
    \midrule
    \rowc \textbf{Normalized MAE} & \textcolor{red}{\textbf{0.544}} & \textcolor{red}{0.553} & 0.729  & 1.041  & 0.657 & 0.598 & 0.576  & 1.000  & 1.028  & 0.927  & 0.758  & 0.872  & 0.898  & 0.785  & 0.760  & 0.741  & 0.759  & 0.783  & 0.749  & 0.770  \\
    \bottomrule
    \end{tabular}%
    }
  \label{tab:monash_full}%
\end{table}%

\subsection{Baselines} \label{subsec:app_baselines}

\paragraph{Baselines} We select multiple representative baselines for comparison, including various time series foundation models as well as other popular TSF baselines covering Transformer-based and MLP-based architectures.
These baseline models selected for comparison are briefly described below:

\begin{enumerate}[leftmargin=*]
    \item \textbf{VisionTS} \citep{VisionTS} is a vision-model-based TSF foundation model which utilizes the visual masked autoencoder pre-trained on ImageNet as the backbone model, and reformulate TSF as a patch-level image reconstruction task to complete prediction.
    
    \item \textbf{Moirai} \citep{Moirai} is an encoder-based TSF foundation model trained on the Large-scale Open Time Series Archive (LOTSA), with over 231B observations across nine domains. It has three variants: \textbf{small}, \textbf{base}, and \textbf{large}.

    \item \textbf{Time-MoE} \citep{Time-MoE} comprises a family of decoder-only transformer models, which leverages a sparse mixture-of-experts (MoE) design by activating only a subset of networks for each prediction to reduce computational load and maintain high model capacity.

    \item \textbf{Chronos} \citep{Chronos} tokenizes time series values using scaling and quantization into a fixed vocabulary, and trains T5 family language models (20M to 710M parameters) on these tokenized time series via the cross-entropy loss.

    \item \textbf{Moment} \citep{Moment} family models serve as a building block for diverse time series analysis tasks, are effective out-of-the-box, and are tunable using in-distribution and task-specific data to improve performance.
    
    \item \textbf{Timer} \citep{Timer} is a decoder-based TSF foundation model exhibiting similar characteristics to LLMs, such as flexible context length and autoregressive generation, along with notable few-shot generalization, scalability, and task generality.
    
    \item \textbf{TimesFM} \citep{TimesFM} is a decoder-style TSF foundation model, using a large time-series corpus comprising both real-world and synthetic datasets.
    
    \item \textbf{LLMTime} \citep{LLMTime} encodes time series data to a text sequence, supporting zero-shot forecasting.
    
    \item \textbf{PatchTST} \citep{PatchTST} uses Transformer encoders with patching and channel independence techniques for improved predictions.

    \item \textbf{TiDE} \citep{TiDE} is an MLP-based encoder-decoder TSF model, which enjoys the simplicity and speed of linear models while also being able to handle covariates and non-linear dependencies.

    \item \textbf{TFT} \citep{TFT} is an attention-based architecture which combines high-performance multi-horizon forecasting with interpretable insights into temporal dynamics.

\end{enumerate}

For the long-term TSF benchmark, we include \visionts and other time series foundation models' results from their individual original papers.
For the Monash and PF benchmark, we include all results from both Moirai and \visionts.
For the GIFT-Eval benchmark, results are obtained from official code repository.

\begin{table}[t]
  \centering
  \caption{Full results of zero-shot forecasting on the long-term TSF benchmark. \textbf{Bold}: the best result.}
  \vspace{-0.5em}
  \resizebox{\linewidth}{!}{
    \begin{tabular}{ccccccccccccccccccccccccccccccccccccccccccc}
    \toprule
    \textbf{Pre-train} &       & \multicolumn{5}{c}{\textit{\textbf{\emoji{figures/em-full_shot} Hybrid}}}    &       & \multicolumn{2}{c}{\textit{\textbf{\emoji{figures/em-frame_with_picture} Images}}}    &       & \multicolumn{32}{c}{{\textbf{\textit{\emoji{figures/em-chart_with_upwards_trend} Time-series}}}} \\
\cmidrule{3-7}\cmidrule{9-10}\cmidrule{12-43}  \textbf{Method} &       & \multicolumn{2}{c}{\textcolor{red}{$\textbf{VisionTS++}_{l}$}} &       & \multicolumn{2}{c}{\textcolor{red}{$\textbf{VisionTS++}_{b}$}} &       & \multicolumn{2}{c}{\textbf{VisionTS}} &       & \multicolumn{2}{c}{$\textbf{Time-MoE}_s$} &       & \multicolumn{2}{c}{$\textbf{Time-MoE}_b$} &       & \multicolumn{2}{c}{$\textbf{Chronos}_s$} &       & \multicolumn{2}{c}{$\textbf{Chronos}_b$} &       & \multicolumn{2}{c}{$\textbf{Chronos}_l$} &       & \multicolumn{2}{c}{$\textbf{Moirai}_s$} &       & \multicolumn{2}{c}{$\textbf{Moirai}_b$} &       & \multicolumn{2}{c}{$\textbf{Moirai}_l$} &       & \multicolumn{2}{c}{\textbf{Moment}} &       & \multicolumn{2}{c}{\textbf{Timer(28B)}} &       & \multicolumn{2}{c}{\textbf{TimesFM}} \\
\cmidrule{3-4}\cmidrule{6-7}\cmidrule{9-10}\cmidrule{12-13}\cmidrule{15-16}\cmidrule{18-19}\cmidrule{21-22}\cmidrule{24-25}\cmidrule{27-28}\cmidrule{30-31}\cmidrule{33-34}\cmidrule{36-37}\cmidrule{39-40}\cmidrule{42-43}    \textbf{Metric} &       & \textbf{MSE} & \textbf{MAE} &       & \textbf{MSE} & \textbf{MAE} &       & \textbf{MSE} & \textbf{MAE} &       & \textbf{MSE} & \textbf{MAE} &       & \textbf{MSE} & \textbf{MAE} &       & \textbf{MSE} & \textbf{MAE} &       & \textbf{MSE} & \textbf{MAE} &       & \textbf{MSE} & \textbf{MAE} &       & \textbf{MSE} & \textbf{MAE} &       & \textbf{MSE} & \textbf{MAE} &       & \textbf{MSE} & \textbf{MAE} &       & \textbf{MSE} & \textbf{MAE} &       & \textbf{MSE} & \textbf{MAE} &       & \textbf{MSE} & \textbf{MAE} \\
    \midrule
    \multirow{5}[2]{*}{\rotatebox{90}{$ETTm1$}} & 96    & 0.312  & \textbf{0.342}  &       & 0.316  & 0.343  &       & 0.341  & 0.347  &       & 0.338  & 0.368  &       & \textbf{0.309 } & 0.357  &       & 0.511  & 0.423  &       & 0.454  & 0.408  &       & 0.457  & 0.403  &       & 0.404  & 0.383  &       & 0.335  & 0.360  &       & 0.353  & 0.363  &       & 0.654  & 0.527  &       & 0.420  & 0.418  &       & 0.361  & 0.370  \\
          & 192   & \textbf{0.341}  & \textbf{0.360}  &       & 0.347  & 0.362  &       & 0.360  & \textbf{0.360} &       & 0.353  & 0.388  &       & 0.346  & 0.381  &       & 0.618  & 0.485  &       & 0.567  & 0.477  &       & 0.530  & 0.450  &       & 0.435  & 0.402  &       & 0.366  & 0.379  &       & 0.376  & 0.380  &       & 0.662  & 0.532  &       & 0.467  & 0.445  &       & 0.414  & 0.405  \\
          & 336   & \textbf{0.361}  & 0.375  &       & 0.368  & 0.379  &       & 0.377  & \textbf{0.374 } &       & 0.381  & 0.413  &       & 0.373  & 0.408  &       & 0.683  & 0.524  &       & 0.662  & 0.525  &       & 0.577  & 0.481  &       & 0.462  & 0.416  &       & 0.391  & 0.394  &       & 0.399  & 0.395  &       & 0.672  & 0.537  &       & 0.502  & 0.467  &       & 0.445  & 0.429  \\
          & 720   & \textbf{0.401}  & \textbf{0.400}  &       & 0.408  & 0.405  &       & 0.416  & 0.405  &       & 0.504  & 0.493  &       & 0.475  & 0.477  &       & 0.748  & 0.566  &       & 0.900  & 0.591  &       & 0.660  & 0.526  &       & 0.490  & 0.437  &       & 0.434  & 0.419  &       & 0.432  & 0.417  &       & 0.692  & 0.551  &       & 0.558  & 0.499  &       & 0.512  & 0.471  \\
          & avg   & \textbf{0.354}  & \textbf{0.369}  &       & 0.360  & 0.372  &       & 0.374  & 0.372  &       & 0.394  & 0.416  &       & 0.376  & 0.406  &       & 0.640  & 0.500  &       & 0.646  & 0.500  &       & 0.556  & 0.465  &       & 0.448  & 0.410  &       & 0.382  & 0.388  &       & 0.390  & 0.389  &       & 0.670  & 0.537  &       & 0.487  & 0.457  &       & 0.433  & 0.419  \\
    \midrule
    \multirow{5}[2]{*}{\rotatebox{90}{$ETTm2$}} & 96    & \textbf{0.167}  & \textbf{0.245}  &       & 0.169  & 0.248  &       & 0.228  & 0.282  &       & 0.201  & 0.291  &       & 0.197  & 0.286  &       & 0.209  & 0.291  &       & 0.199  & 0.274  &       & 0.197  & 0.271  &       & 0.205  & 0.282  &       & 0.195  & 0.269  &       & 0.189  & 0.260  &       & 0.260  & 0.335  &       & 0.247  & 0.324  &       & 0.202  & 0.270  \\
          & 192   & 0.217  & 0.280  &       & \textbf{0.216 } & \textbf{0.279 } &       & 0.262  & 0.305  &       & 0.258  & 0.334  &       & 0.250  & 0.322  &       & 0.280  & 0.341  &       & 0.261  & 0.322  &       & 0.254  & 0.314  &       & 0.261  & 0.318  &       & 0.247  & 0.303  &       & 0.247  & 0.300  &       & 0.289  & 0.350  &       & 0.294  & 0.358  &       & 0.289  & 0.321  \\
          & 336   & 0.261  & 0.311  &       & \textbf{0.260 } & \textbf{0.308 } &       & 0.293  & 0.328  &       & 0.324  & 0.373  &       & 0.337  & 0.375  &       & 0.354  & 0.390  &       & 0.326  & 0.366  &       & 0.313  & 0.353  &       & 0.319  & 0.355  &       & 0.291  & 0.333  &       & 0.295  & 0.334  &       & 0.324  & 0.369  &       & 0.335  & 0.385  &       & 0.360  & 0.366  \\
          & 720   & \textbf{0.329}  & \textbf{0.358}  &       & 0.330  & \textbf{0.358} &       & 0.343  & 0.370  &       & 0.488  & 0.464  &       & 0.480  & 0.461  &       & 0.553  & 0.499  &       & 0.455  & 0.439  &       & 0.416  & 0.415  &       & 0.415  & 0.410  &       & 0.355  & 0.377  &       & 0.372  & 0.386  &       & 0.394  & 0.409  &       & 0.386  & 0.418  &       & 0.462  & 0.430  \\
          & avg   & \textbf{0.244} & \textbf{0.298} &       & \textbf{0.244}  & \textbf{0.298}  &       & 0.318  & 0.366  &       & 0.316  & 0.361  &       & 0.349  & 0.380  &       & 0.310  & 0.350  &       & 0.295  & 0.338  &       & 0.300  & 0.341  &       & 0.272  & 0.321  &       & 0.276  & 0.320  &       & 0.317  & 0.366  &       & 0.316  & 0.371  &       & 0.328  & 0.347  \\
    \midrule
    \multirow{5}[2]{*}{\rotatebox{90}{$ETTh1$}} & 96    & 0.368  & 0.392  &       & 0.369  & 0.392  &       & 0.353  & 0.383  &       & 0.357  & \textbf{0.381 } &       & \textbf{0.350 } & 0.382  &       & 0.466  & 0.409  &       & 0.440  & 0.393  &       & 0.441  & 0.390  &       & 0.375  & 0.402  &       & 0.384  & 0.402  &       & 0.380  & 0.398  &       & 0.688  & 0.557  &       & 0.393  & 0.421  &       & 0.414  & 0.404  \\
          & 192   & 0.401  & 0.412  &       & 0.399  & 0.412  &       & 0.392  & 0.410  &       & \textbf{0.384 } & \textbf{0.404 } &       & 0.388  & 0.412  &       & 0.530  & 0.450  &       & 0.492  & 0.426  &       & 0.502  & 0.424  &       & 0.399  & 0.419  &       & 0.425  & 0.429  &       & 0.440  & 0.434  &       & 0.688  & 0.560  &       & 0.434  & 0.447  &       & 0.465  & 0.434  \\
          & 336   & 0.416  & 0.424  &       & 0.415  & \textbf{0.421 } &       & \textbf{0.407 } & 0.423  &       & 0.411  & 0.434  &       & 0.411  & 0.430  &       & 0.570  & 0.486  &       & 0.550  & 0.462  &       & 0.576  & 0.467  &       & 0.412  & 0.429  &       & 0.456  & 0.450  &       & 0.514  & 0.474  &       & 0.675  & 0.563  &       & 0.460  & 0.464  &       & 0.503  & 0.456  \\
          & 720   & 0.425  & 0.446  &       & 0.424  & 0.437  &       & \textbf{0.406 } & \textbf{0.441 } &       & 0.449  & 0.477  &       & 0.427  & 0.455  &       & 0.615  & 0.543  &       & 0.882  & 0.591  &       & 0.835  & 0.583  &       & 0.413  & 0.444  &       & 0.470  & 0.473  &       & 0.705  & 0.568  &       & 0.683  & 0.585  &       & 0.487  & 0.494  &       & 0.511  & 0.481  \\
          & avg   & 0.403  & 0.418  &       & 0.402  & 0.416  &       & \textbf{0.390 } & \textbf{0.414 } &       & 0.400  & 0.424  &       & 0.394  & 0.420  &       & 0.545  & 0.472  &       & 0.591  & 0.468  &       & 0.589  & 0.466  &       & 0.400  & 0.424  &       & 0.434  & 0.439  &       & 0.510  & 0.469  &       & 0.684  & 0.566  &       & 0.444  & 0.457  &       & 0.473  & 0.444  \\
    \midrule
    \multirow{5}[2]{*}{\rotatebox{90}{$ETTh2$}} & 96    & \textbf{0.267}  & \textbf{0.317}  &       & 0.277  & 0.326 &       & 0.271 & 0.328  &       & 0.305  & 0.359  &       & 0.302  & 0.354  &       & 0.307  & 0.356  &       & 0.308  & 0.343  &       & 0.320  & 0.345  &       & 0.281  & 0.334  &       & 0.277  & 0.327  &       & 0.287  & 0.325  &       & 0.342  & 0.396  &       & 0.308  & 0.369  &       & 0.315  & 0.349  \\
          & 192   & 0.329  & \textbf{0.361}  &       & 0.333  & 0.362  &       & \textbf{0.328 } & 0.367  &       & 0.351  & 0.386  &       & 0.364  & 0.385  &       & 0.376  & 0.401  &       & 0.384  & 0.392  &       & 0.406  & 0.399  &       & 0.340  & 0.373  &       & 0.340  & 0.374  &       & 0.347  & 0.367  &       & 0.354  & 0.402  &       & 0.348  & 0.398  &       & 0.388  & 0.395  \\
          & 336   & 0.350  & \textbf{0.380}  &       & 0.350  & 0.384  &       & \textbf{0.345} & 0.381  &       & 0.391  & 0.418  &       & 0.417  & 0.425  &       & 0.408  & 0.431  &       & 0.429  & 0.430  &       & 0.492  & 0.453  &       & 0.362  & 0.393  &       & 0.371  & 0.401  &       & 0.377  & 0.393  &       & 0.356  & 0.407  &       & 0.366  & 0.414  &       & 0.422  & 0.427  \\
          & 720   & \textbf{0.362}  & \textbf{0.401}  &       & 0.370  & 0.409  &       & 0.388  & 0.422  &       & 0.419  & 0.454  &       & 0.537  & 0.496  &       & 0.604  & 0.533  &       & 0.501  & 0.477  &       & 0.603  & 0.511  &       & 0.380  & 0.416  &       & 0.394  & 0.426  &       & 0.404  & 0.421  &       & 0.395  & 0.434  &       & 0.409  & 0.446  &       & 0.443  & 0.454  \\
          & avg   & \textbf{0.327}  & \textbf{0.365}  &       & 0.333  & 0.370  &       & 0.333  & 0.375  &       & 0.367  & 0.404  &       & 0.405  & 0.415  &       & 0.424  & 0.430  &       & 0.406  & 0.411  &       & 0.455  & 0.427  &       & 0.341  & 0.379  &       & 0.346  & 0.382  &       & 0.354  & 0.377  &       & 0.362  & 0.410  &       & 0.358  & 0.407  &       & 0.392  & 0.406  \\
    \midrule
    \multirow{5}[2]{*}{\rotatebox{90}{$Electricity$}} & 96    & \textbf{0.147}  & 0.233  &       & 0.152  & 0.237  &       & 0.177  & 0.266  &       & - & - &       & - & - &       & 0.157  & 0.234  &       & 0.154  & 0.231  &       & 0.152  & \textbf{0.229 } &       & 0.205  & 0.299  &       & 0.158  & 0.248  &       & 0.152  & 0.242  &       & 0.745  & 0.680  &       & - & - &       & - & - \\
          & 192   & \textbf{0.164}  & \textbf{0.250}  &       & 0.168  & 0.252  &       & 0.188  & 0.277  &       & - & - &       & - & - &       & 0.183  & 0.258  &       & 0.179  & 0.254  &       & 0.172  & 0.252  &       & 0.220  & 0.310  &       & 0.174  & 0.263  &       & 0.171  & 0.259  &       & 0.755  & 0.683  &       & - & - &       & - & - \\
          & 336   & \textbf{0.184}  & \textbf{0.268}  &       & 0.186  & 0.269  &       & 0.207  & 0.296  &       & - & - &       & - & - &       & 0.220  & 0.290  &       & 0.214  & 0.284  &       & 0.203  & 0.276  &       & 0.236  & 0.323  &       & 0.191  & 0.278  &       & 0.192  & 0.278  &       & 0.766  & 0.687  &       & - & - &       & - & - \\
          & 720   & 0.229  & \textbf{0.303}  &       & \textbf{0.228} & \textbf{0.303} &       & 0.256  & 0.337  &       & - & - &       & - & - &       & 0.321  & 0.353  &       & 0.311  & 0.346  &       & 0.289  & 0.337  &       & 0.270  & 0.347  &       & 0.229  & 0.307  &       & 0.236  & 0.313  &       & 0.794  & 0.696  &       & - & - &       & - & - \\
          & avg   & \textbf{0.181}  & \textbf{0.264}  &       & 0.184  & 0.265  &       & 0.207  & 0.294  &       & - & - &       & - & - &       & 0.220  & 0.284  &       & 0.215  & 0.279  &       & 0.204  & 0.274  &       & 0.233  & 0.320  &       & 0.188  & 0.274  &       & 0.188  & 0.273  &       & 0.765  & 0.687  &       & - & - &       & - & - \\
    \midrule
    \multirow{5}[2]{*}{\rotatebox{90}{$Weather$}} & 96    & 0.146  & \textbf{0.179}  &       & \textbf{0.145} & \textbf{0.179} &       & 0.220  & 0.257  &       & 0.160  & 0.214  &       & 0.159  & 0.213  &       & 0.211  & 0.243  &       & 0.203  & 0.238  &       & 0.194  & 0.235  &       & 0.173  & 0.212  &       & 0.167  & 0.203  &       & 0.177  & 0.208  &       & 0.243  & 0.255  &       & 0.243  & 0.283  &       & - & - \\
          & 192   & 0.190  & 0.221  &       & \textbf{0.187 } & \textbf{0.219 } &       & 0.244  & 0.275  &       & 0.210  & 0.260  &       & 0.215  & 0.266  &       & 0.263  & 0.294  &       & 0.256  & 0.290  &       & 0.249  & 0.285  &       & 0.216  & 0.250  &       & 0.209  & 0.241  &       & 0.219  & 0.249  &       & 0.278  & 0.329  &       & 0.288  & 0.320  &       & - & - \\
          & 336   & 0.245  & 0.261  &       & \textbf{0.240 } & \textbf{0.258 } &       & 0.280  & 0.299  &       & 0.274  & 0.309  &       & 0.291  & 0.322  &       & 0.321  & 0.339  &       & 0.314  & 0.336  &       & 0.302  & 0.327  &       & 0.260  & 0.282  &       & 0.256  & 0.276  &       & 0.277  & 0.292  &       & 0.306  & 0.346  &       & 0.323  & 0.345  &       & - & - \\
          & 720   & 0.324  & 0.313  &       & \textbf{0.317 } & \textbf{0.308 } &       & 0.330  & 0.337  &       & 0.418  & 0.405  &       & 0.415  & 0.400  &       & 0.404  & 0.397  &       & 0.397  & 0.396  &       & 0.372  & 0.378  &       & 0.320  & 0.322  &       & 0.321  & 0.323  &       & 0.365  & 0.350  &       & 0.350  & 0.374  &       & 0.362  & 0.374  &       & - & - \\
          & avg   & 0.226  & 0.243  &       & \textbf{0.222 } & \textbf{0.241 } &       & 0.269  & 0.292  &       & 0.266  & 0.297  &       & 0.270  & 0.300  &       & 0.300  & 0.318  &       & 0.293  & 0.315  &       & 0.279  & 0.306  &       & 0.242  & 0.267  &       & 0.238  & 0.261  &       & 0.260  & 0.275  &       & 0.294  & 0.326  &       & 0.304  & 0.331  &       & - & - \\
    \midrule
    \rowc
    \multicolumn{2}{c}{\textbf{Average}} & \textcolor{red}{\textbf{0.289}}  & \textcolor{red}{\textbf{0.326}}  &       & \textcolor{red}{0.291}  & \textcolor{red}{0.327}  &       & 0.309  & 0.345  &       & - & - &       & - & - &       & 0.413  & 0.397  &       & 0.410  & 0.387  &       & 0.396  & 0.379  &       & 0.327  & 0.357  &       & 0.310  & 0.344  &       & 0.329  & 0.350  &       & 0.515  & 0.482  &       & - & - &       & - & - \\
    \rowc
    \multicolumn{2}{c}{\textbf{1st Count}} & \multicolumn{2}{c}{\textcolor{red}{31}} &       & \multicolumn{2}{c}{\textcolor{red}{20}} &       & \multicolumn{2}{c}{9} &       & \multicolumn{2}{c}{3} &       & \multicolumn{2}{c}{2} &       & \multicolumn{2}{c}{0} &       & \multicolumn{2}{c}{0} &       & \multicolumn{2}{c}{1} &       & \multicolumn{2}{c}{0} &       & \multicolumn{2}{c}{0} &       & \multicolumn{2}{c}{0} &       & \multicolumn{2}{c}{0} &       & \multicolumn{2}{c}{0} &       & \multicolumn{2}{c}{0} \\
    \bottomrule
    \end{tabular}%
  }
  \label{tab:ltsf_full}%
\end{table}%

\begin{table}[t]
    \centering
  \begin{minipage}[t]{0.37\textwidth}

  \centering
  \caption{Random initialization (right) vs. Loading MAE pre-trained weights (left) before CPT.}
\vspace{-0.5em}
\resizebox{\linewidth}{!}{
    \begin{tabular}{cccc}
    \toprule
          &       & $\textbf{VisionTS++}_b$ & \textbf{rand\_init} \\
    \midrule
    \textbf{Monash} & MAE & \textbf{0.553 } & 0.733 \\
    \midrule
    \midrule
    \multirow{2}[2]{*}{\textbf{PF}} & MASE  & \textbf{0.677 } & 0.814 \\
          & CRPS  & \textbf{0.515 } & 0.627 \\
    \midrule
    \midrule
    \multirow{2}[2]{*}{\textbf{ETTm1}} & MSE   & \textbf{0.360 } & 0.387 \\
          & MAE   & \textbf{0.372 } & 0.396 \\
    \midrule
    \multirow{2}[2]{*}{\textbf{ETTm2}} & MSE   & \textbf{0.244 } & 0.29 \\
          & MAE   & \textbf{0.298 } & 0.337 \\
    \midrule
    \multirow{2}[2]{*}{\textbf{ETTh1}} & MSE   & \textbf{0.402 } & 0.447 \\
          & MAE   & \textbf{0.416 } & 0.45 \\
    \midrule
    \multirow{2}[2]{*}{\textbf{ETTh2}} & MSE   & \textbf{0.333 } & 0.47 \\
          & MAE   & \textbf{0.370 } & 0.439 \\
    \midrule
    \multirow{2}[2]{*}{\textbf{Electricity}} & MSE   & \textbf{0.184 } & 0.225 \\
          & MAE   & \textbf{0.265 } & 0.298 \\
    \midrule
    \multirow{2}[2]{*}{\textbf{Weather}} & MSE   & \textbf{0.222 } & 0.233 \\
          & MAE   & \textbf{0.241 } & 0.257 \\
    \bottomrule
    \end{tabular}%
  }
  \label{tab:ablation_rand_init}%
    \end{minipage}
    \hfill
    \begin{minipage}[t]{0.59\textwidth}
        \centering
  \caption{Ablation studies on each component in the  \model.}
\vspace{-0.5em}

  \resizebox{\columnwidth}{!}{
  
    \begin{tabular}{cccccc}
    \toprule
          &       & $\textbf{VisionTS++}_b$ & \textbf{w/o quantile} & \textbf{w/o filter} & \textbf{w/o color} \\
    \midrule
    \textbf{Monash} & MAE   & \textbf{0.553 } & 0.593  & 0.578  & 0.634  \\
    \midrule
    \midrule
    \multirow{2}[2]{*}{\textbf{PF}} & MASE  & \textbf{0.677 } & 0.714  & 0.690  & 0.725  \\
          & CRPS  & \textbf{0.515 } & 0.551  & 0.531  & 0.565  \\
    \midrule
    \midrule
    \multirow{2}[2]{*}{\textbf{ETTm1}} & MSE   & \textbf{0.360 } & 0.392  & 0.388  & 0.408  \\
          & MAE   & \textbf{0.372 } & 0.401  & 0.397  & 0.419  \\
    \midrule
    \multirow{2}[2]{*}{\textbf{ETTm2}} & MSE   & \textbf{0.244 } & 0.278  & 0.270  & 0.302  \\
          & MAE   & \textbf{0.298 } & 0.328  & 0.324  & 0.356  \\
    \midrule
    \multirow{2}[2]{*}{\textbf{ETTh1}} & MSE   & \textbf{0.402 } & 0.421  & 0.416  & 0.453  \\
          & MAE   & \textbf{0.416 } & 0.438  & 0.425  & 0.464  \\
    \midrule
    \multirow{2}[2]{*}{\textbf{ETTh2}} & MSE   & \textbf{0.333 } & 0.355  & 0.336  & 0.376  \\
          & MAE   & \textbf{0.370 } & 0.387  & 0.372  & 0.402  \\
    \midrule
    \multirow{2}[2]{*}{\textbf{Electricity}} & MSE   & \textbf{0.184 } & 0.208  & 0.189  & 0.215  \\
          & MAE   & \textbf{0.265 } & 0.288  & 0.272  & 0.299  \\
    \midrule
    \multirow{2}[2]{*}{\textbf{Weather}} & MSE   & \textbf{0.222 } & 0.234  & 0.228  & 0.245  \\
          & MAE   & \textbf{0.241 } & 0.259  & 0.249  & 0.271  \\
    \bottomrule
    \end{tabular}%
  }
  \label{tab:ablation_all}%
    \end{minipage}
\end{table}

\section{Full Experimental Results}
\subsection{Full Results for In-distribution Monash Benchmark} \label{subsec:app_monash_full}

\par Table \ref{tab:monash_full} provides the full breakdown of results ffor the Monash benchmark, listing results for each dataset in Monash. Based on the table, \model not only obtains SOTA overall normalized MAE results, but also achieves the best results in the vast majority of cases.

\subsection{Full Results for Out-of-distribution LTSF Benchmark} \label{subsec:app_ltsf_full}
\par Table \ref{tab:ltsf_full} provides the full detailed results for the long-term time series forecasting experiments, listing results for each prediction length. 
From the results, we can see that \model~ achieves the best results in most cases (large: 31 out of 62, and base: 20 out of 62), outperforming \visionts~ (9 out of 62), Time-MoE (3 out of 62), and all other models.

\subsection{Full results for Random Initialization and Ablation Study} \label{subsec:app_ablation_full}
\par We report the results of random initialization of \model in Table \ref{tab:ablation_rand_init}, and the results of ablation studies in Table \ref{tab:ablation_all} due to space limit. Analysis of these experiment results are detailed in Section \ref{subsec:further_analysis} in the full text.

\section{Visualization} \label{sec:app_visual}
\par
\begin{figure*}[t]
  \centering
  \subfigure[Input Image]{
    \includegraphics[width=0.23\textwidth]{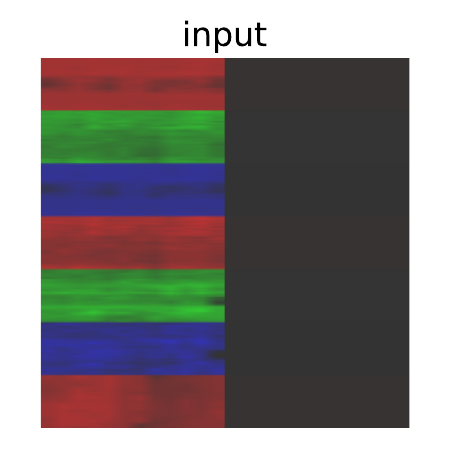}
  }\quad\quad\quad\quad
  \subfigure[Reconstructed Image]{
    \includegraphics[width=0.23\textwidth]{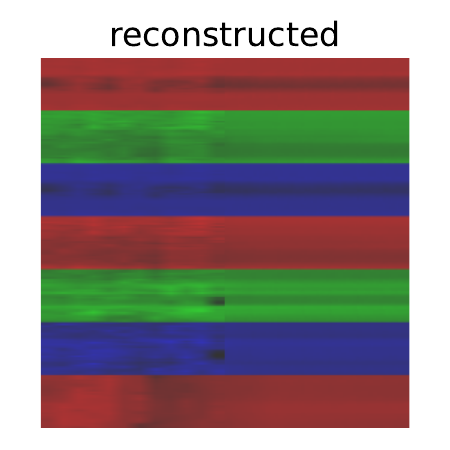}
  }
  \vspace{-1.1em}
  \subfigure[\model]{
    \includegraphics[width=0.97\textwidth]{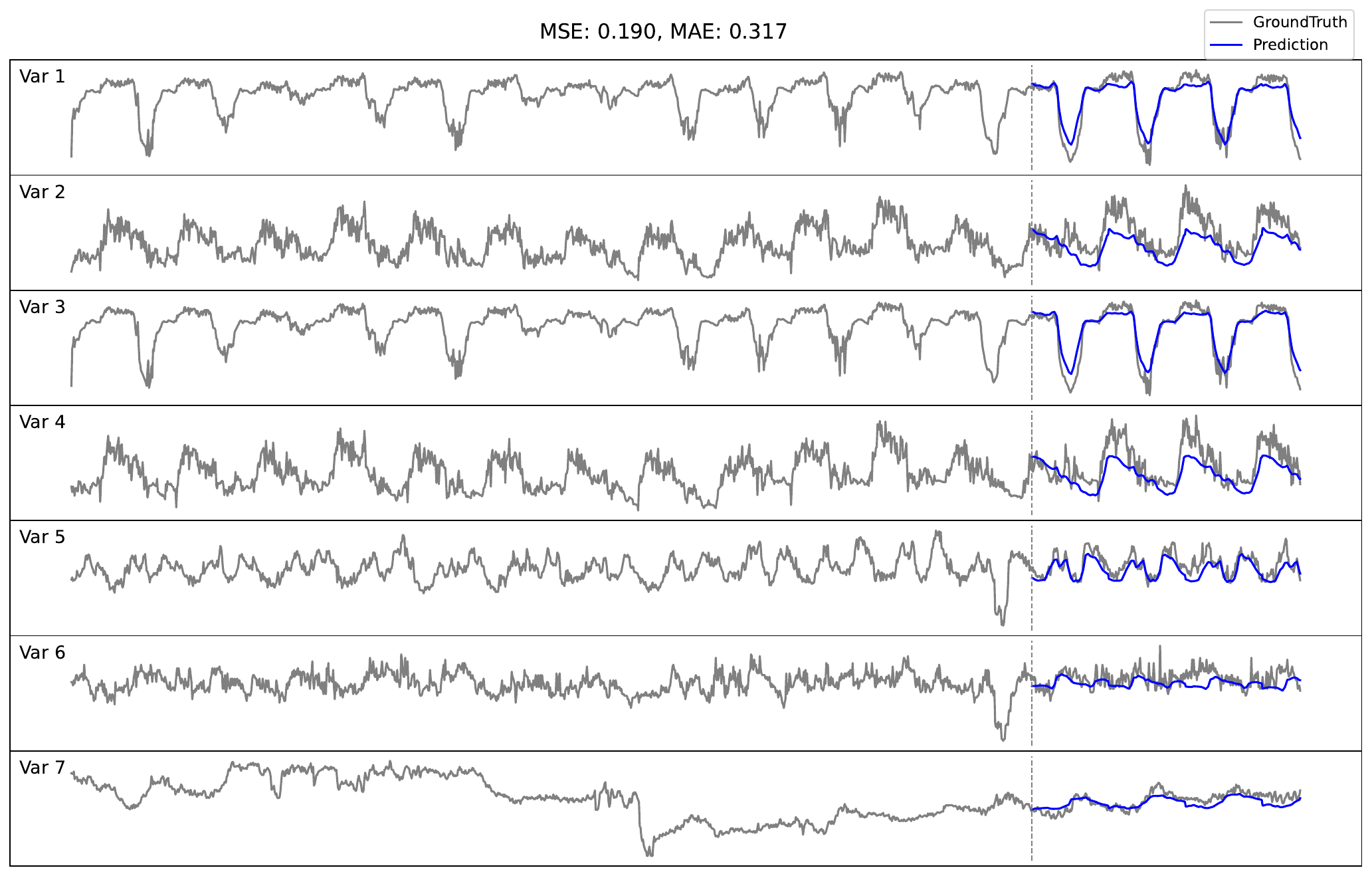}
  }
  \subfigure[\visionts]{
    \includegraphics[width=0.97\textwidth]{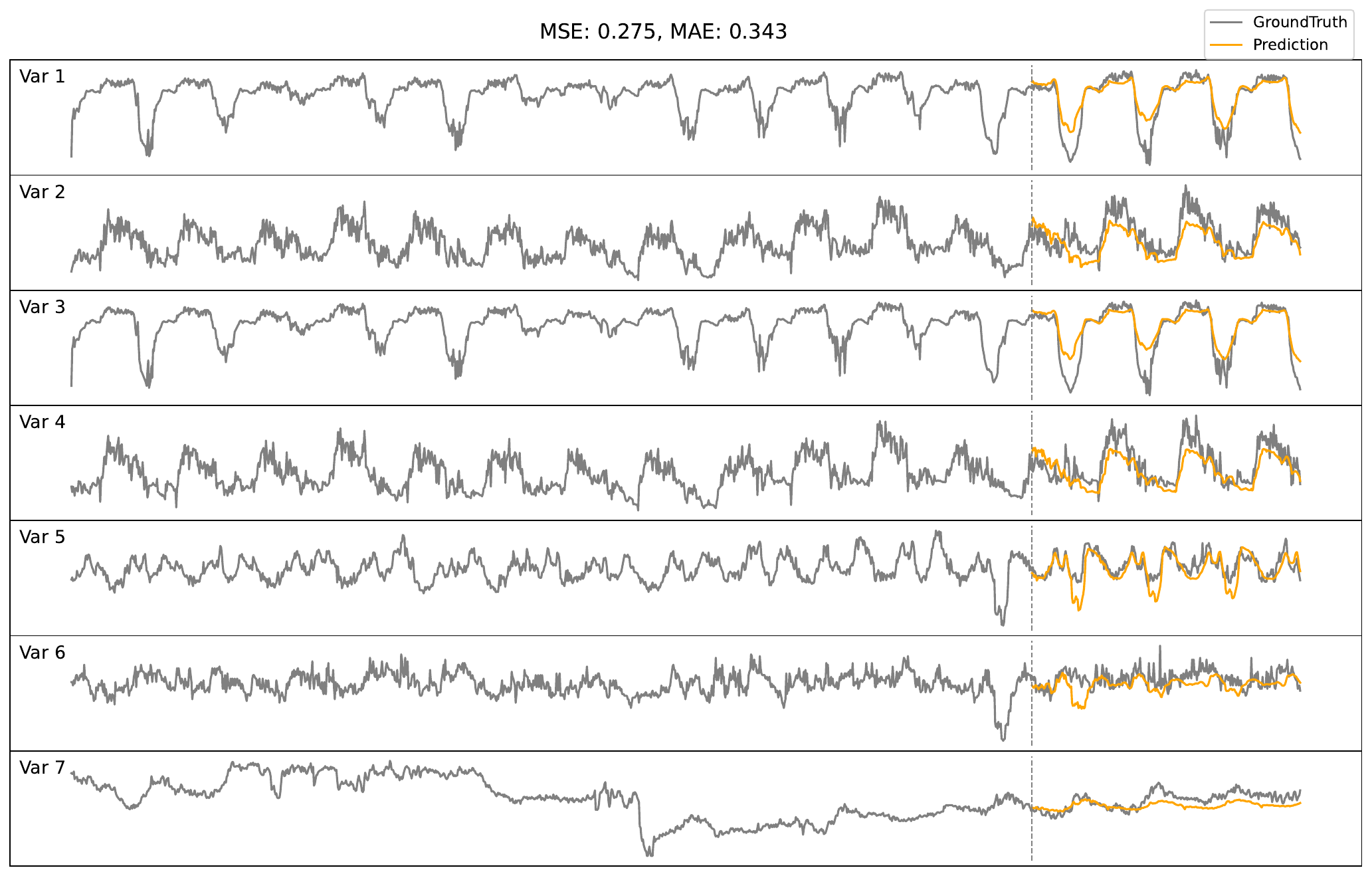}
  }
  \vspace{-0.8em}
  \caption{Forecasting visualization on a sample from ETTm1. (a-b) Input/Output images of \model. (c-d) Prediction comparison between \model and \visionts.}
  \label{fig:ettm1_case1}
\end{figure*}

\begin{figure*}[t]
  \centering
  \subfigure[Input Image]{
    \includegraphics[width=0.23\textwidth]{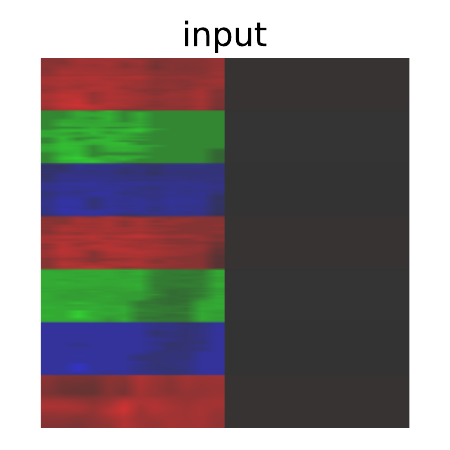}
  }\quad\quad\quad\quad
  \subfigure[Reconstructed Image]{
    \includegraphics[width=0.23\textwidth]{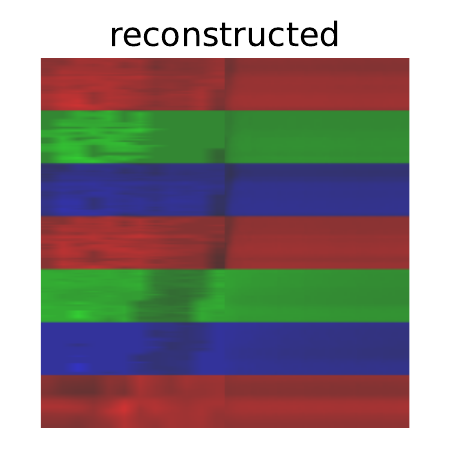}
  }
  \vspace{-1.1em}
  \subfigure[\model]{
    \includegraphics[width=0.97\textwidth]{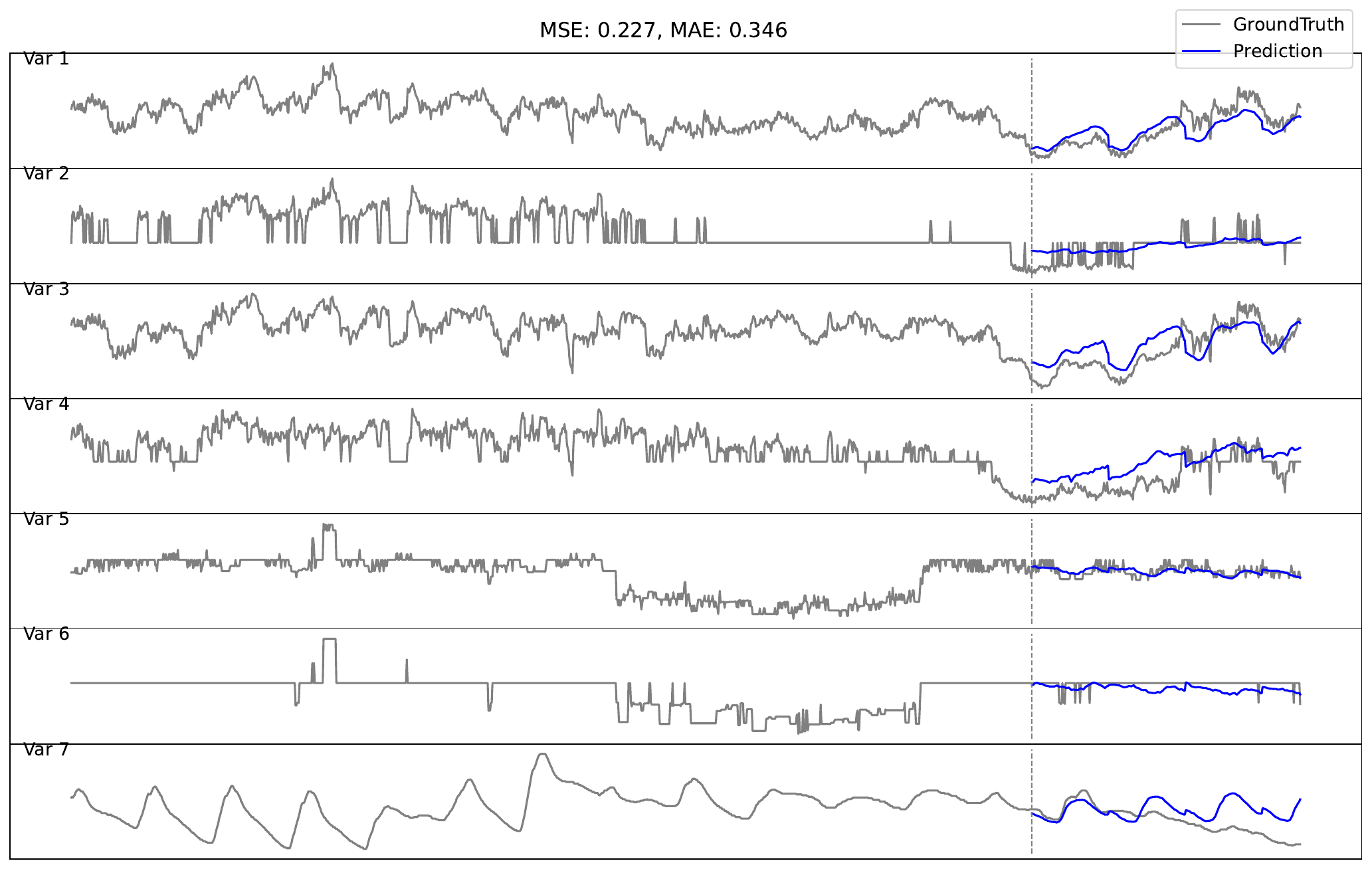}
  }
  \subfigure[\visionts]{
    \includegraphics[width=0.97\textwidth]{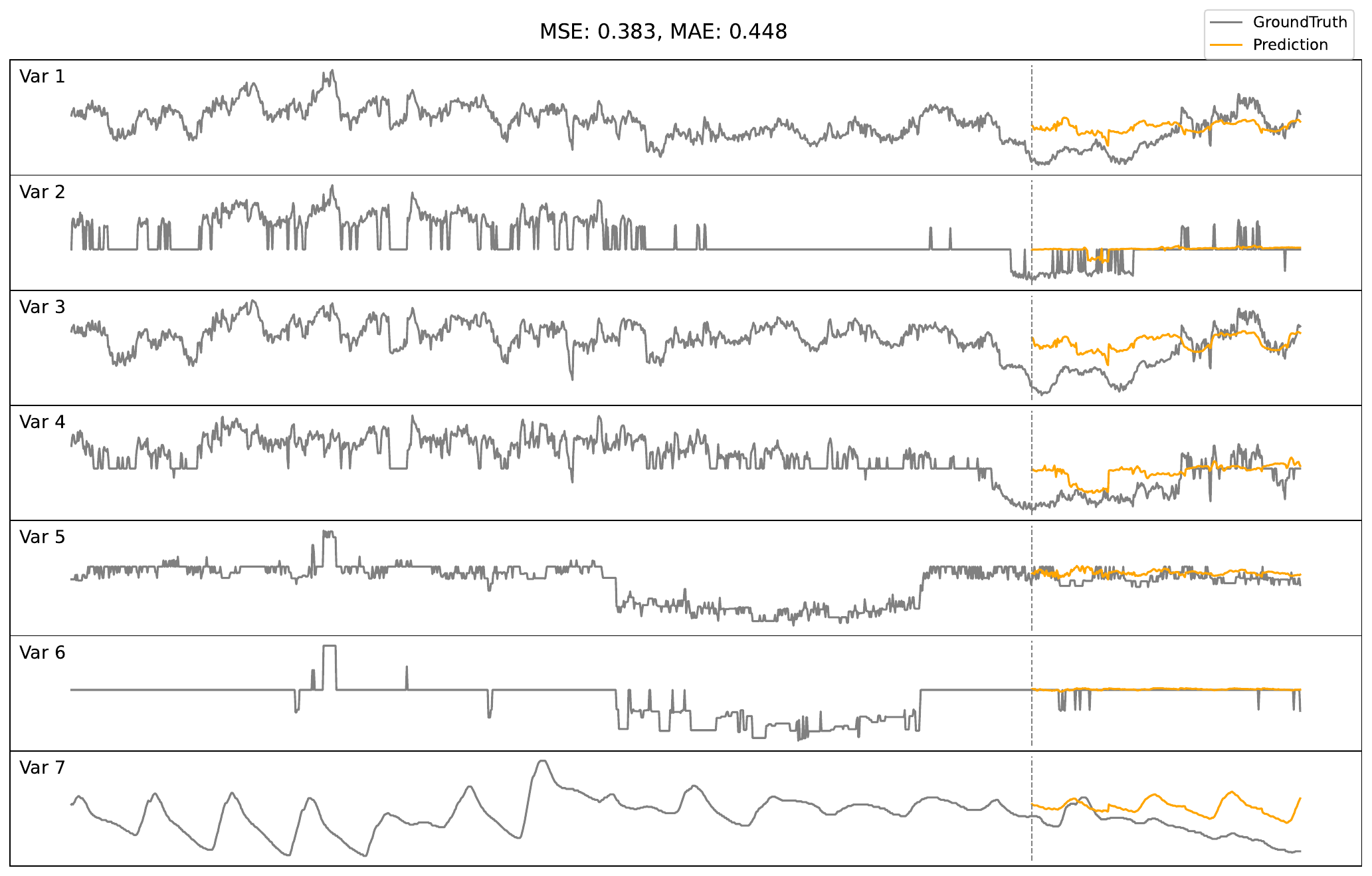}
  }
  \vspace{-0.8em}
  \caption{Forecasting visualization on a sample from ETTm2. (a-b) Input/reconstructed images of \model. (c-d) Prediction comparison between \model and \visionts.}
  \label{fig:ettm2_case1}
\end{figure*}

 In this section, we visualize the multivariate time series predictions of \model in the zero-shot setting, including its input and reconstructed images. 
We also visualize its predictions, with MSE and MAE metrics for comparison. These samples are presented in Figure \ref{fig:ettm1_case1} and Figure \ref{fig:ettm2_case1}.

\par These examples show the superior forecasting performance of \model over \visionts after conducting the continual pre-training, as well as other components that effectively address the modality gaps between images and time series.

\end{document}